\pdfoutput=1
\documentclass[conference]{IEEEtran}
\usepackage{amsmath,amssymb,amsfonts}
\usepackage{array}
\usepackage{booktabs}
\usepackage{graphicx}
\usepackage{cite}
\usepackage{tabularx}
\usepackage{float}
\usepackage{multirow}
\usepackage{subcaption}
\usepackage{url}
\graphicspath{{figures/}}

\begin{document}

\title{Physics-Enhanced Deep Learning for Proactive Thermal Runaway Forecasting in Li-Ion Batteries}

\author{
Salman Khan,
Syed Sajid Ullah*, 
Muhammad Zunair Zamir,
Jie Li,\\
Abdul~Malik,
and Saeed~Mian~Qaisar
}
\maketitle

\begin{abstract}
Accurate prediction of thermal runaway in lithium-ion batteries is essential for ensuring the safety, efficiency, and reliability of modern energy storage systems. Conventional data-driven approaches, such as Long Short-Term Memory (LSTM) networks, can capture complex temporal dependencies but often violate thermodynamic principles, resulting in physically inconsistent predictions. Conversely, physics-based thermal models provide interpretability but are computationally expensive and difficult to parameterize for real-time applications. To bridge this gap, this study proposes a Physics-Informed Long Short-Term Memory (PI-LSTM) framework that integrates governing heat transfer equations directly into the deep learning architecture through a physics-based regularization term in the loss function. The model leverages multi-feature input sequences including state of charge, voltage, current, mechanical stress, and surface temperature to forecast battery temperature evolution while enforcing thermal diffusion constraints. Extensive experiments conducted on thirteen lithium-ion battery datasets demonstrate that the proposed PI-LSTM achieves a remarkable 81.9\% reduction in root mean square error (RMSE) and 81.3\% lower mean absolute error (MAE) compared to the standard LSTM baseline, while also outperforming CNN--LSTM and multilayer perceptron (MLP) models by wide margins. The inclusion of physical constraints enhances the model's generalization across diverse operating conditions and eliminates non-physical temperature oscillations. These results confirm that physics-informed deep learning offers a viable pathway toward interpretable, accurate, and real-time thermal management in next-generation battery systems.
\end{abstract}


% -----

\section{Introduction}
Lithium-ion batteries (LIBs) serve as the backbone of modern electrification, powering electric vehicles (EVs), consumer electronics, and renewable energy systems. Their high energy density and long cycle life make them a leading energy storage technology. However, under abusive conditions, such as overcharging, overheating, internal short circuits, or mechanical deformation, LIBs are susceptible to \textit{thermal runaway} (TR), a chain reaction where exothermic reactions accelerate uncontrollably, leading to fire, explosion, or toxic gas release \cite{Gu2025StrainPerspective, Li2023Impact, Tang2024NailPenetration}. Predicting TR events in real-time is crucial for battery management systems (BMSs) aiming to prevent catastrophic failure.

Conventional TR prediction techniques rely on deterministic, physics-based thermal models that simulate heat generation and transfer within the battery system \cite{Kong2023Review, Wang2023Multiscale}. These models, grounded in the laws of thermodynamics and electrochemical kinetics, offer physical interpretability and are useful in controlled environments \cite{Kong2022_CFD}. However, their applicability in real-world, dynamic settings is limited due to high computational demands, sensitivity to parameterization, and the need for internal measurements that may not be accessible during operation \cite{Liu2023InternalShort, Li2023Multifunctional}.

Conversely, data-driven models such as neural networks and ensemble learning algorithms have shown promise in extracting complex temporal dependencies from sensor data streams, enabling near real-time prediction of anomalous events \cite{Li2022DLPrognosis, Sun2022Online, Jiang2021RealData}. Recurrent neural networks (RNNs), particularly Long Short-Term Memory (LSTM) networks, are widely adopted due to their ability to model sequential dependencies in multivariate time series. Despite their success, pure deep learning models often generate outputs that violate physical laws, e.g., predicting negative temperatures or oscillatory thermal responses, because they lack intrinsic understanding of thermodynamic constraints \cite{Chen2024Hybrid, Cheng2024Charging}.

To address these shortcomings, recent research has explored hybrid frameworks that combine physical models with data-driven learning. Such methods include using impedance spectroscopy \cite{Dong2021Impedance, Lyu2021Impedance}, gas sensors \cite{Song2025GasSensor, Das2023Gas}, and embedded internal sensors \cite{Kisseler2024InternalSensor, Han2023NDIR} to detect precursors of thermal instability. However, these methods often depend on specialized hardware or post hoc rule enforcement, limiting their scalability and generalizability across diverse battery chemistries and form factors.

This study introduces a Physics-Informed Long Short-Term Memory (PI-LSTM) framework that embeds thermal diffusion equations directly into the learning architecture through a regularization term in the loss function. The model utilizes inputs such as current, voltage, temperature, mechanical stress, and state of charge, ensuring both accuracy and thermodynamic consistency. Experiments on 13 real-world datasets show that the proposed approach not only surpasses traditional LSTM, CNN--LSTM, and MLP models but also eliminates non-physical temperature behavior while maintaining high generalization across varying usage patterns.

This study makes the following contributions:
\begin{itemize}
    \item Proposes a physics-informed long short-term memory (PI-LSTM) framework that integrates the heat diffusion equation into the learning process, enabling physically consistent and interpretable temperature forecasting.
    \item Demonstrates enhanced predictive accuracy and stability compared with conventional data-driven architectures, effectively reducing non-physical fluctuations across diverse cells and operating conditions.
    \item Conducts a systematic ablation analysis on the influence of physics-based regularization and discusses its implications for efficient, real-time deployment in battery management systems.
\end{itemize}

The remainder of this paper is organized as follows: Section~\ref{sec:relatedwork} reviews related work on thermal runaway prediction, highlighting key advancements and research gaps. Section~\ref{sec:methodology} details the proposed Physics-Informed LSTM (PI-LSTM) framework, including the model architecture, physics-based regularization, and training process. Section~\ref{sec:results} presents the experimental setup, benchmark comparisons, and in-depth analysis of the results. Finally, Section~\ref{sec:conclusion} concludes the study and outlines potential directions for future research.

\section{Related Work}
\label{sec:relatedwork}
Research in thermal runaway forecasting spans a rich spectrum, including analytical models, sensor-based early detection systems, and data-driven machine learning techniques.

Physics-based models are traditionally used to simulate heat generation, conduction, and dissipation in LIBs under operational and abusive conditions \cite{Kong2023Review}. These include conjugate heat transfer models \cite{Kong2022_CFD}, multiphysics finite element simulations \cite{Wang2023Multiscale}, and impact-based damage analysis of multifunctional sandwich structures \cite{Li2023Multifunctional}. While accurate and interpretable, these models are computationally intensive and require extensive calibration for each battery configuration \cite{Liu2023InternalShort, Li2023Impact}.

Sensor-based methods utilize signals such as impedance, gas evolution, internal temperature, or pressure to issue early warnings. Electrochemical impedance spectroscopy (EIS) is used for in-situ monitoring of cell degradation and thermal precursors \cite{Dong2021Impedance, Lyu2021Impedance}. Gas sensors based on photoacoustic spectroscopy \cite{Liao2022Photoacoustic}, NDIR detection \cite{Han2023NDIR}, and off-gas detection \cite{Purushothaman2024OffGas, Song2025GasSensor} provide real-time alerts for chemical instability. Internal sensor networks offer high-fidelity temperature readings from the core of the battery pack \cite{Kisseler2024InternalSensor}. However, these systems can be cost-prohibitive or invasive for commercial BMS integration.

Machine learning and deep learning algorithms have been increasingly applied to TR forecasting using real-world vehicle data \cite{Jiang2021RealData, Jia2023RiskEV}. LSTM models \cite{Li2022DLPrognosis, Sun2022Online}, DBSCAN clustering for anomaly detection \cite{Li2019DBSCAN}, and entropy-based predictors \cite{Hong2021Entropy} have shown good performance in identifying early fault signatures. Ensemble methods that integrate voltage, temperature, and current signals into data-driven models have achieved high accuracy in predictive maintenance tasks \cite{Jiang2021Hybrid, Guo2023HealthScore}.

To improve model interpretability and robustness, hybrid frameworks have been introduced that couple machine learning with physical domain knowledge. Chen et al. \cite{Chen2024Hybrid} combined neural networks with thermal laws for TR warning. Klink et al. \cite{Klink2022Comparison} contrasted model-based versus sensor-based systems and advocated for hybridization. Despite their promise, many such models lack direct embedding of physics into training loss and often require rule-based post-processing.

\begin{table}[h]
\caption{Summary of representative methods for thermal runaway prediction.}
\label{tab:tr_summary_methods}
\centering
\setlength{\tabcolsep}{2pt}
\renewcommand{\arraystretch}{1.05}
\begin{tabularx}{\columnwidth}{@{}c p{2.0cm} p{2.9cm} p{2.9cm}@{}}
\toprule
\textbf{Ref.} & \textbf{Methodology} & \textbf{Key Features} & \textbf{Limitations} \\
\midrule
\cite{Gu2025StrainPerspective} & Strain-based early warning & Utilizes mechanical strain sensors for early thermal detection. & Limited to strain sensors; lacks thermal modeling. \\
\cite{Wang2023Multiscale} & Multiscale CFD model & Multi-scale CFD simulation including particle venting dynamics. & Too slow for real-time use. \\
\cite{Li2022DLPrognosis} & Deep learning (LSTM) with abnormal heat input & Deep LSTM network with temperature and heat features for prediction. & No physical constraints; prone to oscillations. \\
\cite{Sun2022Online} & Online LSTM with multi-sensor inputs & Real-time sensor fusion using LSTM and voltage/current features. & Overfits when generalizing to new chemistries. \\
\cite{Jiang2021RealData} & Real-world vehicle ML model & ML model trained on real-world vehicle dataset with feature selection. & Not transferable to new battery packs. \\
\cite{Chen2024Hybrid} & Data-driven + thermal model hybrid & Combines thermal models with deep learning for accuracy and interpretability. & Thermal model is simplified. \\
\cite{Cheng2024Charging} & Charging-phase ML warning & Charging-specific data-driven approach; focused on thermal rise detection. & Only valid during charging stage. \\
\cite{Dong2021Impedance} & EIS-based monitoring & Electrochemical impedance-based degradation sensing approach. & Impedance noise in aging cells. \\
\cite{Lyu2021Impedance} & Impedance-based real-time model & Applies impedance spectroscopy in real time; adaptive signal tracking. & Sensor drift; not robust under fast charging. \\
\cite{Han2023NDIR} & NDIR-based sensor array & NDIR gas sensing system for early gas emission detection. & Sensor calibration needed; cost issues. \\
\cite{Li2019DBSCAN} & DBSCAN anomaly clustering & Unsupervised clustering of multivariate battery signals using DBSCAN. & Static thresholds and no physics guidance. \\
\cite{Gu2025SOS} & State of safety estimator with hybrid logic & State of safety index with hybrid rule learning for early warning. & Requires rule-based safety models. \\
\bottomrule
\end{tabularx}
\end{table}

As summarized in Table~\ref{tab:tr_summary_methods}, the existing body of literature on thermal runaway prediction, while extensive, exhibits several key limitations. Physics-based models, though accurate, are often computationally prohibitive for real-time deployment. Sensor-driven techniques typically depend on costly or intrusive instrumentation, limiting their practicality in large-scale systems. Purely data-driven approaches may generate predictions that contradict fundamental physical laws, reducing reliability and interpretability. Moreover, many hybrid strategies enforce physical constraints only after model training, rather than integrating physics directly into the learning process. To address these challenges, this work embeds thermal diffusion principles directly within an LSTM architecture, enabling real-time forecasting while ensuring adherence to physical consistency. To the best of our knowledge, this study represents the first large-scale deployment of a physics-informed LSTM (PI-LSTM) framework for proactive thermal runaway forecasting across diverse lithium-ion battery datasets.

% ------- Layout -----

% 1. Experimental Setup
% 2. Mechanical Abuse and 
% Structural Response

% Different Modes (Compression, axial, nail, etc.)

% 3. Parametric and Sensitivity Analysis

% Analysis (SOC analysis, radius analysis, position analysis, etc)

% Thesis Topic: 2.2.2 Position and SOC on the safety performance of lithium batteries
% Fig 12 Battery surface temperature evolution diagram under different SOCs

% 2.2.3 Effect of indentation location on the safety performance of lithium batteries

% 2.2.4 Effect of indenter size on the safety performance of lithium batteries

% 3.7.1 Analysis of simulation results under cylindrical indenter
% (1) Analysis of simulation results with different SOC 
% (2) Analysis of the simulation results at different indentation positions 
% (3) Analysis of simulation results at different speeds 

% 3.7.2 Analysis of simulation results under different indenter shapes

% 4. Severity Index

% 5.LSTM-Based Temperature Prediction
% 6. Physics-Informed LSTM-Based Temperature Prediction
% 7. Model Performance

% ----- End Layout ----------

\section{Experimental Framework and Model Development}
\label{sec:methodology}

\subsection{Experimental Setup}

\begin{figure}[H]
\centering
\includegraphics[width=0.48\textwidth]{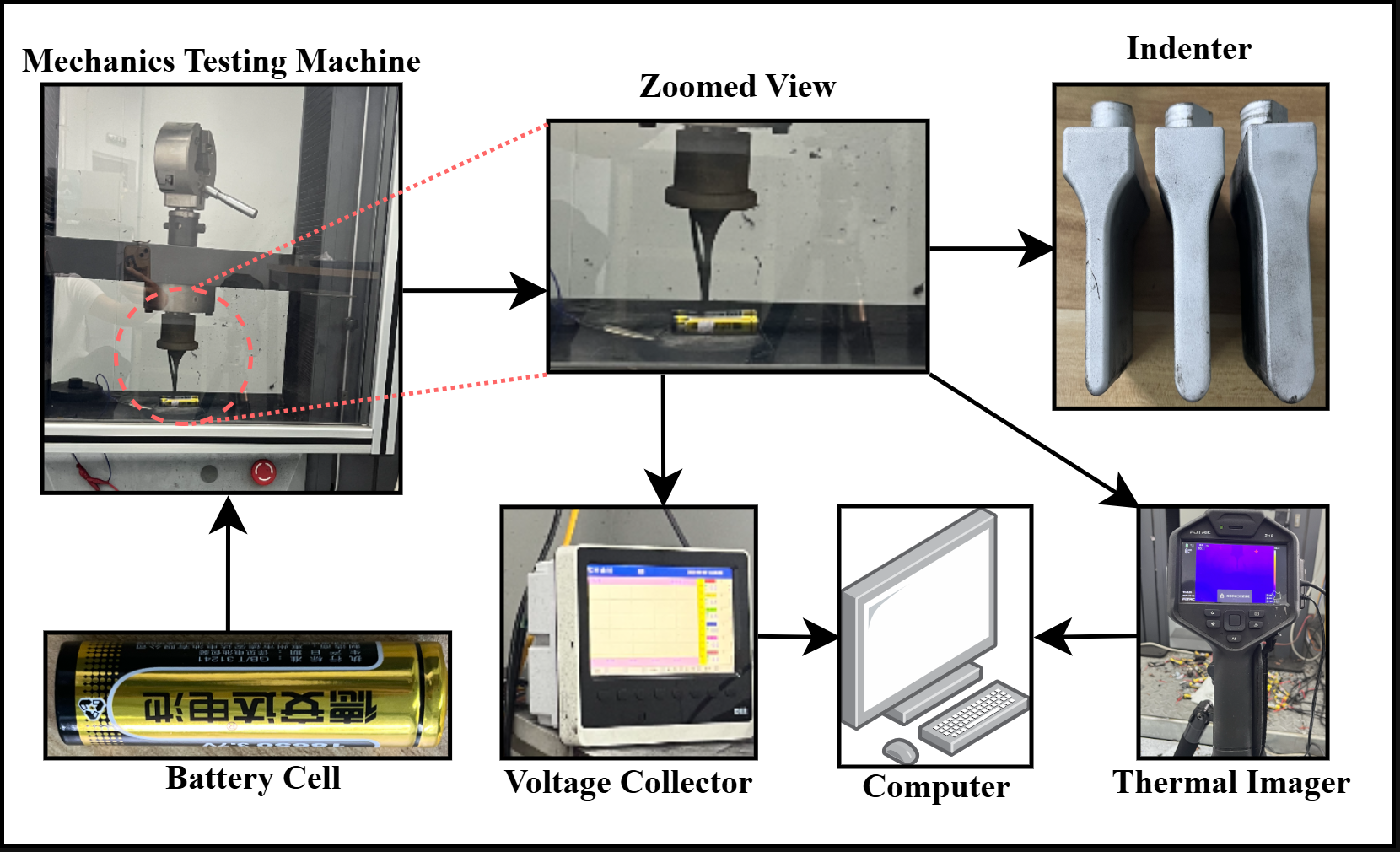}
\caption{Experimental Setup for Thermal Runaway}
\label{fig:battery_damage}
\end{figure}

The experimental platform employed in this study is illustrated in Fig.~\ref{fig:battery_damage}. A commercial 18650-type cylindrical lithium-ion battery (LIB) was selected as the test subject. The battery features a graphite anode and a lithium cobalt oxide (LiCoO$_2$) cathode, with a nominal energy capacity of 4600~mWh. Structurally, the cell consists of a steel casing that encloses a multilayer jellyroll composed of wound anode, cathode, and separator layers. The cell dimensions are 18~mm in diameter and 65~mm in height.

To simulate mechanical abuse conditions and observe the onset of thermal runaway (TR), a multi-instrument experimental setup was constructed. A mechanics testing machine (Fig.~\ref{fig:battery_damage}) is used to apply mechanical loading to the battery with a maximum force capacity of 100~kN. The loading head interfaces with various types of indenters (flat, cylindrical, spherical), as shown in the top-right section of Fig.~\ref{fig:battery_damage}, enabling different abuse scenarios such as indentation, radial and axial compression, and nail penetration.

During mechanical loading, the voltage response of the LIB is recorded in real-time using a voltage collector. This data is transmitted to a computer for analysis. Concurrently, a thermal imager is positioned 1~m from the LIB to capture the surface temperature distribution and maximum temperature rise. The imager is calibrated with an emissivity value of 0.95, verified via thermocouple measurements at the battery's geometric center. Thermal images are captured at 1-second intervals to monitor the thermal behavior throughout the test.

This integrated platform enables synchronized acquisition of mechanical, electrical, and thermal data, which are critical for characterizing the TR behavior of LIBs under abusive mechanical conditions. The experimental procedure includes four test modes: (1) indentation with cylindrical and spherical indenters, (2) radial compression, (3) axial compression, and (4) nail penetration.

\subsection{Mechanical Abuse and Structural Response}

\textbf{Indentation with Cylindrical and Spherical Indenters.}
Indentation tests using cylindrical and spherical indenters are employed to impose localized mechanical intrusion on the battery surface in a controlled manner. Progressive penetration leads to internal layer deformation, separator fracture, and eventual internal short circuits, allowing detailed analysis of staged mechanical-electrical-thermal failure behavior.

\textbf{Radial Compression.}
Radial compression applies a lateral load perpendicular to the battery axis, simulating large-area squeezing during side-impact or pack-level deformation. This mode induces collapse of internal components over a broad region, potentially causing internal short circuits and thermal runaway under distributed mechanical stress.

\textbf{Axial Compression.}
Axial compression subjects the battery to compressive loading along its longitudinal axis, representing crushing forces during structural failure or collision. The resulting deformation of electrode and separator layers can trigger internal short circuits, enabling investigation of thermomechanical coupling under axial intrusion.

\textbf{Nail Penetration.}
Nail penetration involves direct piercing of the battery using a rigid steel needle, producing an immediate and highly localized internal short circuit. This aggressive abuse mode leads to rapid voltage drop and intense localized heating, making it a critical test for evaluating thermal runaway initiation and early warning performance.

\subsection{Severity Index}

Under nail penetration, the battery exhibits the highest failure severity, characterized by an abrupt voltage collapse and a rapid temperature escalation exceeding 350~$^\circ$C, accompanied by violent gas release and sustained thermal runaway. Experimental results show that thermal runaway is triggered almost immediately after internal short-circuit formation, with minimal delay between penetration and peak temperature, indicating an extremely aggressive failure mode.

For axial compression, the severity is comparatively moderate, with internal short circuits occurring after progressive structural deformation. The temperature rise is slower than that observed under nail penetration, and the peak surface temperature typically remains below 300~$^\circ$C. Thermal runaway onset is delayed, reflecting a more gradual failure evolution governed by layer collapse and separator fracture along the cell axis.

In the case of radial compression, the severity index is lower than that of axial compression, as the applied load is distributed over a larger contact area. Experimental observations indicate delayed voltage drop and a slower temperature increase, with peak temperatures generally below those recorded in axial compression tests. Thermal runaway, when it occurs, develops over an extended time scale, demonstrating a less aggressive but still hazardous failure progression.

\subsection{Mechanical Abuse Response Analysis}

This section presents a systematic analysis of the mechanical abuse responses of cylindrical lithium-ion cells, focusing on the coupled mechanical, electrical, and thermal behaviors observed under different loading conditions. Experimental results are organized to isolate the effects of axial position, loading rate, contact geometry, and state of charge on deformation, internal short-circuit initiation, voltage collapse, and thermal escalation. Rather than treating these responses independently, the analysis emphasizes their interdependence, highlighting how mechanical damage governs electrical failure and subsequent thermal runaway. This response-oriented organization provides a unified framework for comparing abuse severity across conditions and establishes physically interpretable trends that support correlation analysis and physics-informed learning in subsequent sections.

% -------------------------------------------------
\subsubsection{Position-Dependent Response}

% The mechanical response of the cylindrical lithium-ion cell under indentation shows a clear dependence on the axial loading position. As shown in Fig.~\ref{fig:pos_force_time} and Fig.~\ref{fig:pos_force_disp}, the indentation force increases quasi-linearly with time and displacement at all positions, followed by a distinct peak associated with the onset of internal damage. The magnitude of this peak force varies systematically with axial location, with mid-height loading exhibiting the highest resistance and delayed post-peak softening, while positions closer to the cell ends show reduced peak force and earlier degradation. This positional trend is summarized in Fig.~\ref{fig:pos_peak_force}, which demonstrates a monotonic decrease in peak force with increasing distance from the mechanically robust mid-height region.

\begin{figure}[H]
    \centering
    \begin{subfigure}[t]{0.48\columnwidth}
        \centering
        \includegraphics[width=0.95\linewidth]{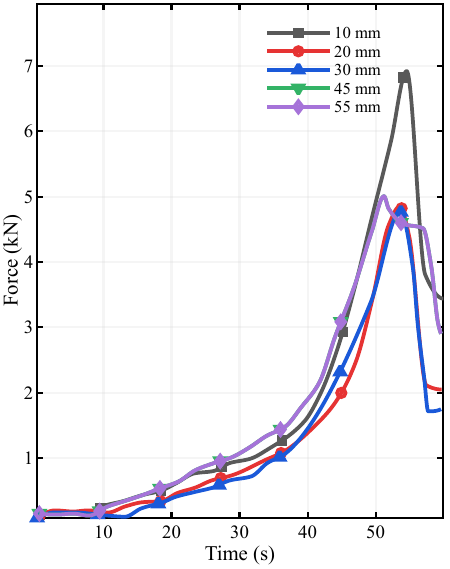}
        \caption{Force--time response at different axial positions}
        \label{fig:pos_force_timea}
    \end{subfigure}
    \hfill
    \begin{subfigure}[t]{0.48\columnwidth}
        \centering
        \includegraphics[width=\linewidth]{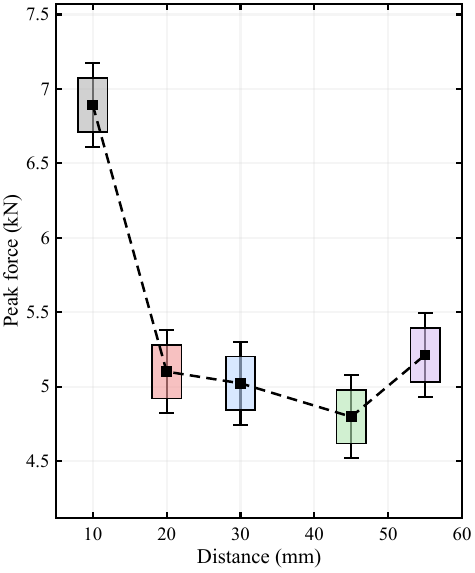}
        \caption{Force--displacement response at different axial positions}
        \label{fig:pos_force_dispb}
    \end{subfigure}

    \caption{Position-dependent mechanical response of the cylindrical lithium-ion cell under indentation loading.}
    \label{fig:pos_mech_response}
\end{figure}

The indentation behavior reflects the mechanical response of the cylindrical lithium-ion cell and shows strong axial position dependence. As shown in Fig.~\ref{fig:pos_mech_response}, all tested locations initially exhibit a quasi-linear force increase due to elastic--plastic deformation of the steel casing and compaction of the jelly-roll. Fig.~\ref{fig:pos_force_timea} reveals that indentations near the cell ends experience reduced peak force ($F_{\text{peak}}$) and earlier softening, caused by geometric discontinuities and packing heterogeneity, whereas mid-height positions demonstrate the highest $F_{\text{peak}}$ owing to symmetric constraint of the electrode layers. This trend is quantified in Fig.~\ref{fig:pos_force_dispb}, which displays a monotonic decline in $F_{\text{peak}}$ with increasing distance from mid-height, along with greater data scatter near the ends---indicative of heightened local structural variability.

The position-dependent peak force is modeled by an exponential decay law:
\begin{equation}
    F_{\text{peak}}(x^*) = F_{\max} \exp\!\left(-\lambda_{\text{pos}} \, x^*\right),
\end{equation}
where $x^*$ is the normalized axial distance from the mechanically optimal location (e.g., mid-height), $F_{\max}$ is the maximum attainable peak force, and $\lambda_{\text{pos}}$ quantifies the decay rate of structural integrity along the cell axis---capturing the gradual loss of confinement and load-bearing capacity near geometric or material discontinuities.

\begin{figure}[H]
    \centering
    \begin{subfigure}[t]{0.48\columnwidth}
        \centering
        \includegraphics[width=\linewidth]{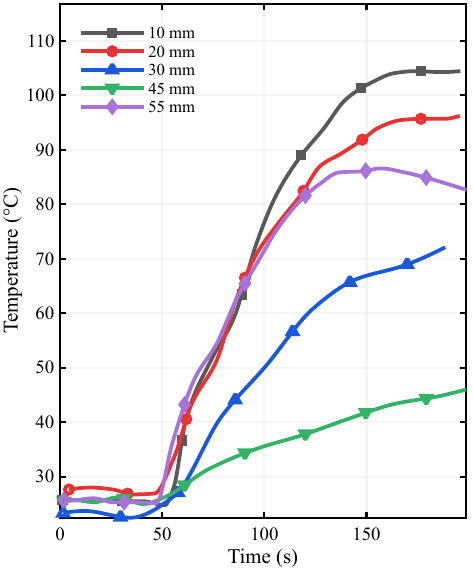}
        \caption{Temperature--time response at different axial positions}
        \label{fig:pos_temp_time}
    \end{subfigure}
    \hfill
    \begin{subfigure}[t]{0.48\columnwidth}
        \centering
        \includegraphics[width=\linewidth]{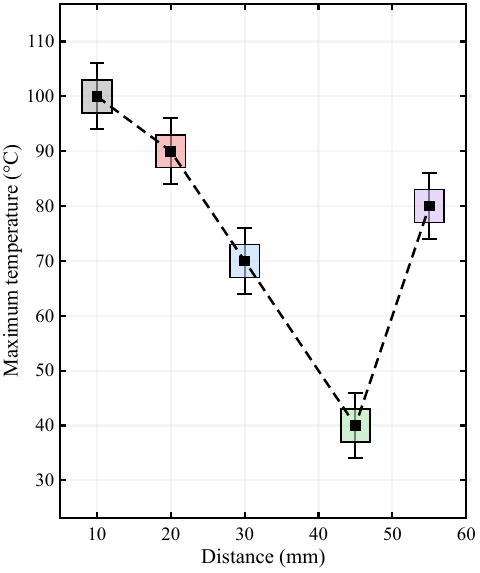}
        \caption{Temperature rise as a function of displacement at different axial positions}
        \label{fig:pos_temp_disp}
    \end{subfigure}

    \caption{Position-dependent thermal response of the cylindrical lithium-ion cell under indentation loading.}
    \label{fig:pos_thermal_response}
\end{figure}

Fig.~\ref{fig:pos_thermal_response} reveals strong coupling between indentation position and thermal evolution, during early loading, temperature remains nearly constant across all positions (Fig.~\ref{fig:pos_temp_time}), indicating negligible heat generation before internal damage. A sharp temperature rise occurs only after short-circuit onset, confirming Joule heating, not mechanical dissipation, as the dominant thermal driver. Positions with lower peak force exhibit earlier, steeper heating (Fig.~\ref{fig:pos_temp_disp}), while stronger regions show delayed, milder temperature rise---evidence that reduced structural resistance accelerates separator failure and electro-thermal runaway at lower mechanical energy input.

The observed trend supports an energy-controlled heating mechanism:
\begin{equation}
    T_{\text{peak}} \approx T_0 + \alpha \, W, \qquad 
    W = \int_{0}^{u_f} F(u) \, du,
\end{equation}
where $W$ is the external mechanical work to failure (area under the force--displacement curve), $T_0$ is the initial surface temperature, and $\alpha$ is an effective thermal conversion coefficient. Positions with smaller $W$ reach thermal runaway more rapidly, and often at higher heating rates, due to localized current concentration during early short-circuit formation.

\begin{figure}[h]
    \centering
    \includegraphics[width=1\columnwidth]{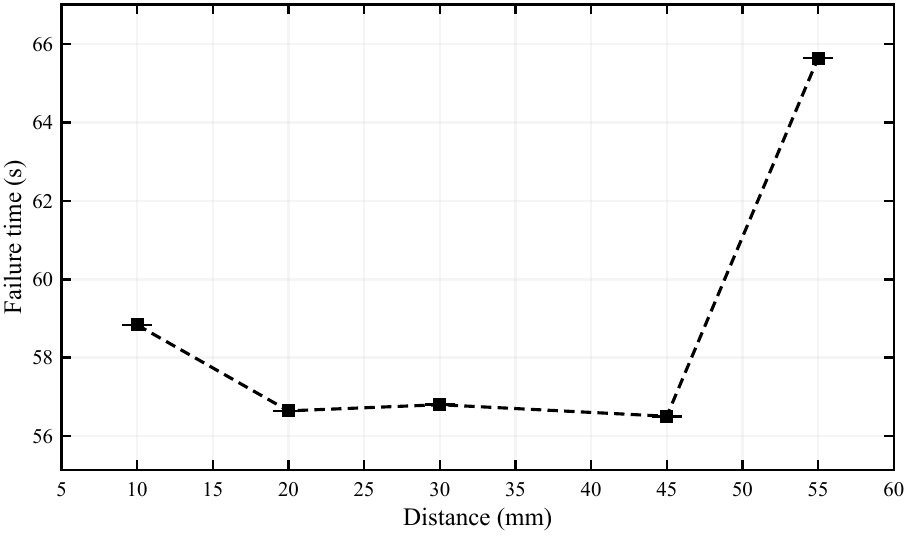}
    \caption{Failure time and corresponding failure distance of the cylindrical lithium-ion cell under indentation at different axial positions.}
    \label{fig:pos_failure_time_distance}
\end{figure}

The electrical response of the cell exhibits strong axial position dependence, as shown in Fig.~\ref{fig:pos_failure_time_distance}: voltage collapse time is inversely correlated with mechanical resistance. Indentations at structurally weaker positions trigger earlier electrical failure (i.e., rapid voltage drop), whereas mechanically robust regions, such as near mid-height, delay the onset of voltage collapse, reflecting greater resilience to internal short-circuit formation.

The voltage collapse time follows a phenomenological exponential trend:
\begin{equation}
    t_{\text{collapse}}(x^*) = t_0 + C \exp\!\left(\lambda_{\text{pos}} \, x^*\right),
\end{equation}
where $x^*$ is the normalized axial distance from the mechanically optimal location, $t_0$ is the baseline collapse time (e.g., at mid-height), $C$ scales the positional sensitivity, and $\lambda_{\text{pos}} > 0$ reflects how rapidly structural weakening accelerates electrical failure. The increasing $t_{\text{collapse}}$ with decreasing $x^*$ confirms that higher mechanical resistance delays separator breakdown and short-circuit propagation, highlighting that electrical failure is governed by mechanically induced internal short formation, not displacement alone.

\subsubsection{Loading-rate (indenter speed) effect}

\begin{figure}[H]
    \centering
    \includegraphics[width=1\columnwidth]{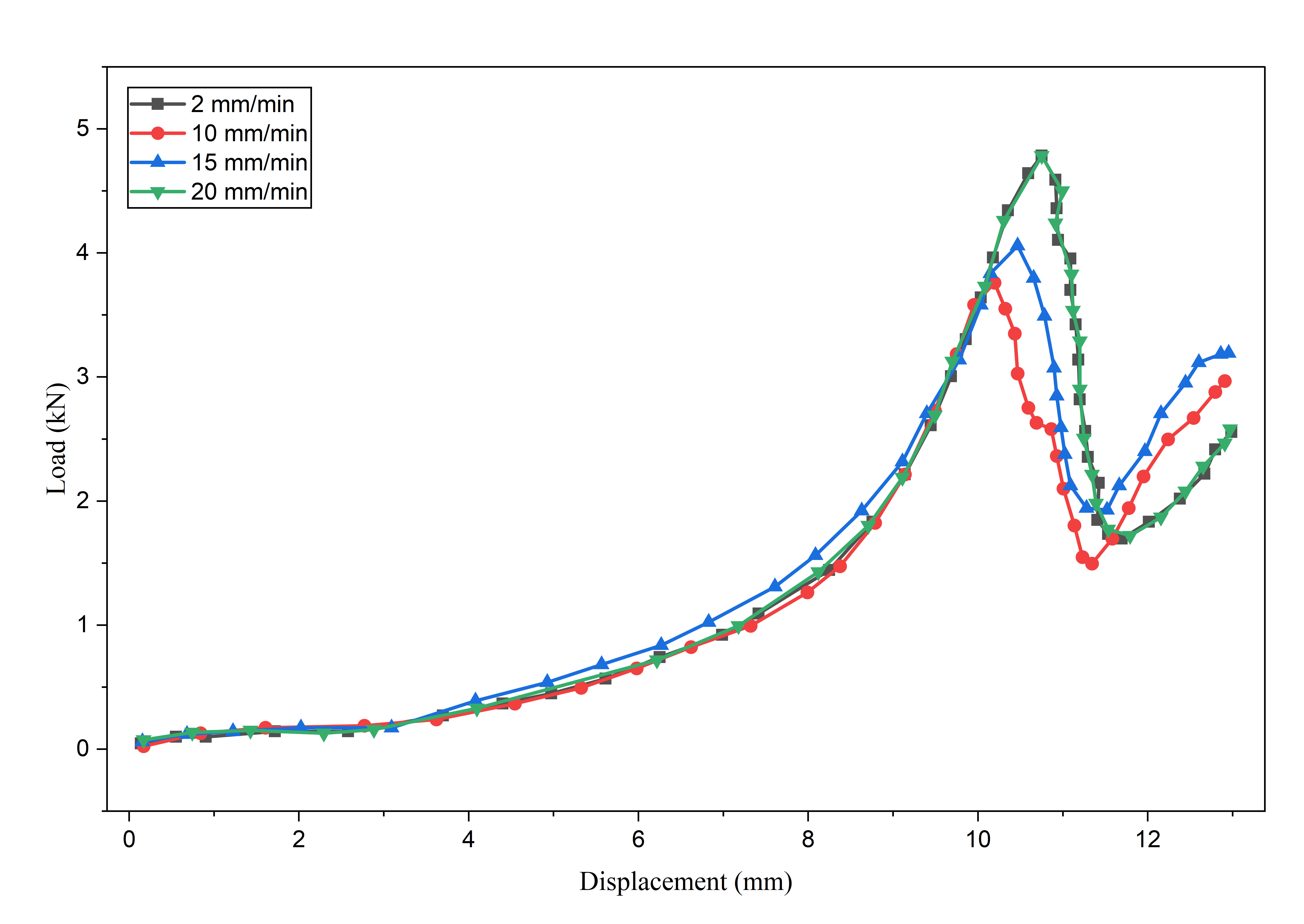}
    \caption{Mechanical response of the cylindrical lithium-ion cell under indentation loading, illustrating the influence of indenter speed on the force--displacement behavior.}
    \label{fig:speed_force_disp}
\end{figure}

As shown in Fig.~\ref{fig:speed_force_disp}, the pre-peak loading is highly repeatable across indentation speeds, with peak force tightly clustered at $\sim$4.76--4.79\,kN, indicating weak rate sensitivity of the global structural resistance (18650 casing + jelly-roll) in maximum load. In contrast, post-peak behavior is speed-dependent: after $F_{\text{peak}}$, curves diverge due to rate-sensitive damage evolution, faster deformation localizes failure at high rates, while slower rates enable gradual energy dissipation through layer compaction and frictional sliding, altering softening and recovery characteristics.

A standard phenomenological representation of the weak rate sensitivity of peak force is given by the logarithmic law:
\begin{equation}
    F_{\text{peak}}(v) = F_0 + k \, \ln\!\left(\frac{v}{v_0}\right),
\end{equation}
where $F_0$ is the reference peak force at reference speed $v_0$, and $k$ is a small coefficient (visually near-zero in the data, reflecting minimal variation in $F_{\text{peak}}$ across tested speeds).

\begin{figure}[H]
    \centering
    \includegraphics[width=1\columnwidth]{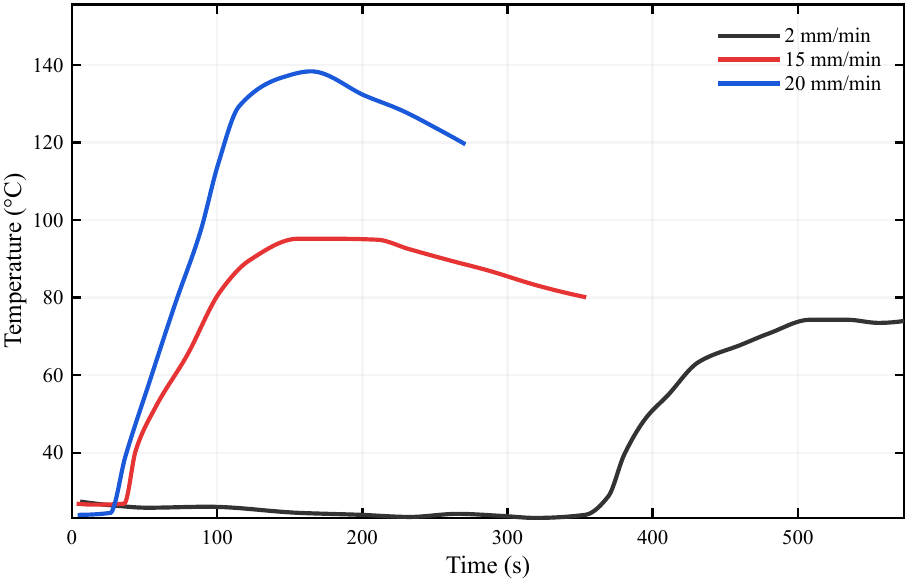}
    \caption{Thermal response of the cylindrical lithium-ion cell under indentation loading, showing the effect of indenter speed on the temperature--time evolution.}
    \label{fig:speed_temp_time}
\end{figure}

\begin{figure}[H]
    \centering
    \includegraphics[width=1\columnwidth]{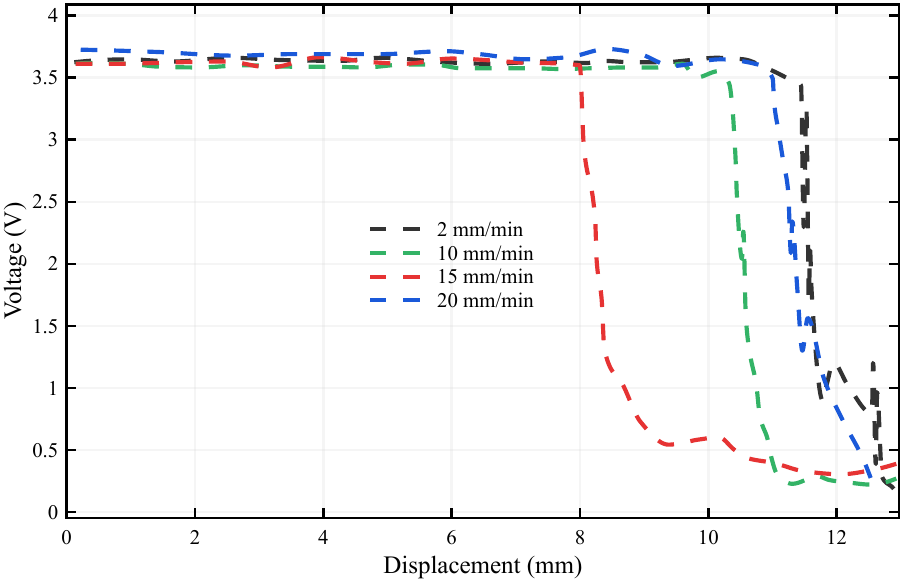}
    \caption{Voltage response as a function of indenter displacement for different indenter speeds during indentation loading of the cylindrical lithium-ion cell.}
    \label{fig:voltage_displacement_speed}
\end{figure}

Fig.~\ref{fig:voltage_displacement_speed} shows that indenter speed critically governs thermal response: at low speed (2~mm/min), rises slowly to a modest plateau---indicating delayed/limited short-circuit formation and effective heat dissipation; in contrast, at higher speeds (15--20~mm/min), rapid temperature spikes occur early, signaling prompt internal short initiation and dominant Joule heating that outpaces conduction, yielding higher peak temperatures.

The energy-rate competition is captured by  
\begin{equation}
    \Delta T_{\text{peak}} \equiv T_{\text{peak}} - T_0 \approx \eta \, W_f, \qquad 
    W_f = \int_0^{u_f} F(u)\,du,
\end{equation}
with an effective conversion efficiency $\eta(v) \approx \eta_0 + c \ln\!\left(\frac{v}{v_0}\right)$ (empirical), where $\eta$ increases with speed due to reduced heat loss, making the failure more adiabatic at higher rates.

Fig.~\ref{fig:voltage_displacement_speed} shows voltage remains near OCV during early deformation and collapses abruptly upon internal short formation; critically, the collapse displacement $u_{\text{collapse}}$ is speed-dependent: lower-speed tests often exhibit larger $u_{\text{collapse}}$ (delayed shorting), while higher speeds can trigger earlier collapse at smaller $u$, reflecting rate-sensitive separator failure and deformation localization.

A flexible yet physics-compatible fit for the speed-dependent collapse displacement is:
\begin{equation}
    u_{\text{collapse}}(v) = a + b \ln v + c (\ln v)^2,
\end{equation}
capturing U- or inverted-U-shaped trends (e.g., competing progressive damage vs. rapid localization). The non-monotonicity reflects that short formation can occur earlier at intermediate speeds, consistent with Joule heating accelerating local breakdown once contact is established.

% -------------------------------------------------
\subsubsection{Indenter Geometry Effect}

\begin{figure}[h]
    \centering
    \begin{subfigure}[t]{0.48\columnwidth}
        \centering
        \includegraphics[width=\linewidth]{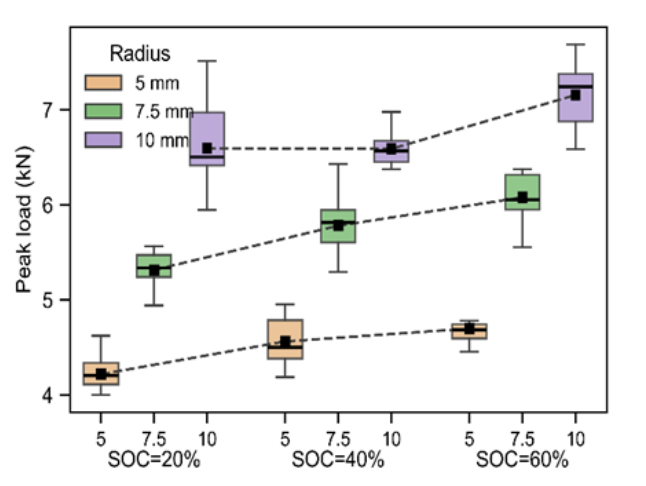}
        \caption{Peak indentation force as a function of indenter radius and state of charge (SOC)}
        \label{fig:geom_force}
    \end{subfigure}
    \hfill
    \begin{subfigure}[t]{0.48\columnwidth}
        \centering
        \includegraphics[width=\linewidth]{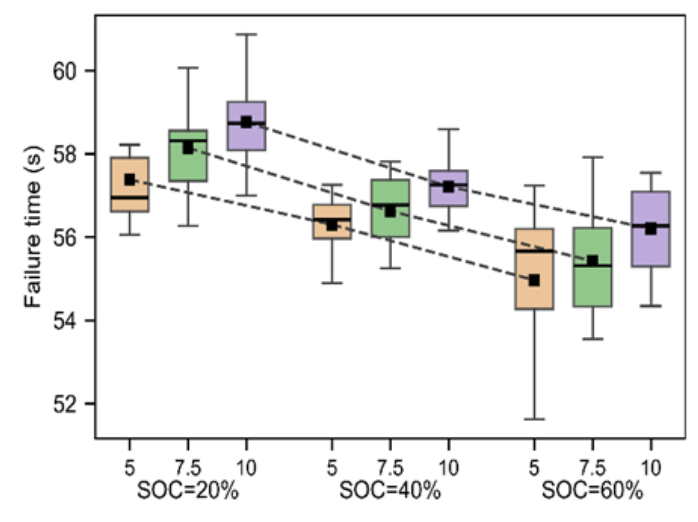}
        \caption{Electrical failure time as a function of state of charge (SOC) under different contact conditions}
        \label{fig:geom_failure_time}
    \end{subfigure}

    \caption{Influence of indenter geometry and state of charge on the mechanical resistance and electrical failure behavior of the cylindrical lithium-ion cell under indentation loading.}
    \label{fig:geometry_soc_response}
\end{figure}

The mechanical response under indentation exhibits a strong dependence on indenter radius and state of charge (SOC). As shown in Fig.~\ref{fig:geom_force}, the peak indentation force increases systematically with increasing indenter radius across all tested SOC levels. For a fixed SOC, larger-radius indenters consistently produce higher peak loads, reflecting the reduction in local contact stress concentration with increasing radius and the need for higher global force to induce comparable levels of local deformation and internal damage. In addition, higher SOC cells exhibit larger peak forces for all radii, indicating that the electrochemical state influences the effective mechanical resistance of the jelly-roll, likely through changes in internal pressure and electrode stiffness. The observed dependence of peak force on indenter radius can be described by an exponential-type structural resistance relation or, within the limited radius range investigated, by a linear approximation capturing the gradual increase in load-bearing capacity as stress localization decreases.

The dependence of peak indentation force on indenter radius can be accurately described by an exponential-type structural resistance relation,
\begin{equation}
F_{\mathrm{peak}}(r) = F_{\max}\exp\!\left(-\lambda_{r} r\right),
\end{equation}
or, equivalently, within the limited indenter radius range investigated, by a linear approximation,
\begin{equation}
F_{\mathrm{peak}}(r) \approx F_{0} + k_{r} r,
\end{equation}
where $r$ denotes the indenter radius and $k_{r} > 0$ captures the gradual increase in load-bearing capacity as stress localization decreases.

The electrical response, characterized by the time to failure, shows a weaker but consistent dependence on indenter radius. As illustrated in Fig.~\ref{fig:geom_failure_time}, failure time decreases with increasing SOC for all contact conditions, while the influence of indenter radius remains secondary but non-negligible. Larger-radius indenters promote a broader deformation zone, facilitating the formation of distributed internal contact regions and accelerating the initiation of internal short circuits despite higher mechanical resistance. At lower SOC, failure times remain comparatively long across all radii, whereas at higher SOC levels, failure occurs earlier for all contact geometries, confirming that stored electrochemical energy amplifies susceptibility to mechanically triggered electrical failure.

At lower state of charge (SOC = 20\%), failure times are generally longer, whereas at higher SOC levels (40--60\%) the failure time shortens across all indenter radii. This behavior confirms that increased stored electrochemical energy amplifies the susceptibility of the cell to mechanically triggered internal short circuits.

This trend may be expressed phenomenologically as
\begin{equation}
t_{\mathrm{fail}}(r) = t_{0} - \beta r,
\end{equation}
where $\beta$ is small but positive, indicating a weak inverse dependence of electrical failure time on indenter radius.

\begin{figure}[H]
    \centering
    \begin{subfigure}[t]{0.48\columnwidth}
        \centering
        \includegraphics[width=\linewidth]{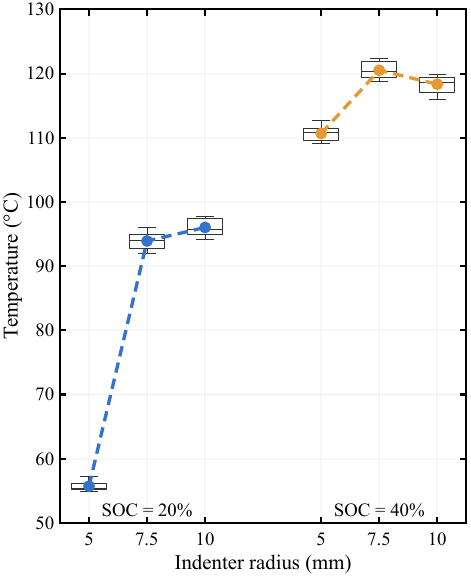}
        \caption{Peak temperature as a function of indenter radius}
        \label{fig:thermal_radius}
    \end{subfigure}
    \hfill
    \begin{subfigure}[t]{0.48\columnwidth}
        \centering
        \includegraphics[width=\linewidth]{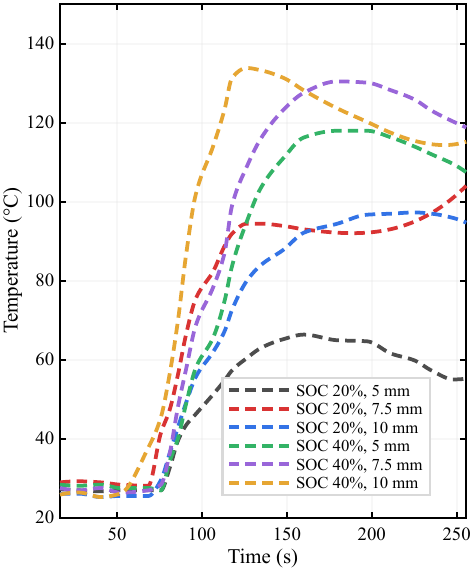}
        \caption{Temperature--time response at different states of charge (SOC)}
        \label{fig:thermal_soc_time}
    \end{subfigure}

    \caption{Thermal response of the cylindrical lithium-ion cell under indentation loading as a function of indenter geometry and state of charge.}
    \label{fig:thermal_geometry_soc}
\end{figure}

The thermal response exhibits the strongest sensitivity to both indenter radius and SOC. As shown in Fig.~\ref{fig:thermal_radius}, the peak temperature increases monotonically with indenter radius, with larger-radius indenters producing substantially higher maximum temperatures. The corresponding temperature--time histories in Fig.~\ref{fig:thermal_soc_time} further show that higher SOC cells experience faster temperature rise rates and earlier thermal escalation. This behavior indicates strong coupling between contact geometry and electrochemical energy availability. Larger-radius indenters generate greater total mechanical work prior to failure, which is efficiently converted into Joule heating following internal short formation. Moreover, the larger contact area promotes more extensive short-circuit paths, increasing electrical dissipation and elevating surface temperature.

The observed trend supports an energy-based thermal scaling,
\begin{equation}
T_{\mathrm{peak}} = T_{0} + \alpha W,
\qquad
W = \int_{0}^{u_{f}} F(u)\,\mathrm{d}u,
\end{equation}
where $T_{0}$ is the baseline temperature prior to failure, $\alpha$ is an energy-to-temperature conversion coefficient, and $W$ denotes the total mechanical work performed up to the onset of failure. Larger-radius indenters generate greater mechanical work prior to failure, which is efficiently converted into Joule heating following internal short formation. Moreover, larger contact areas promote more extensive short-circuit paths, and the resulting electrical dissipation elevates the surface temperature.

Taken together, the mechanical peak load trends (Fig.~\ref{fig:geom_force}), electrical failure characteristics (Fig.~\ref{fig:geom_failure_time}), and thermal responses (Fig.~\ref{fig:thermal_radius} and Fig.~\ref{fig:thermal_soc_time}) demonstrate that indenter radius plays a critical role in governing mechanical abuse severity. Larger-radius indenters increase global mechanical resistance while simultaneously promoting broader internal damage zones, leading to earlier electrical failure and substantially higher thermal output. This coupled behavior underscores the importance of contact geometry in mechanical abuse scenarios and supports the inclusion of indenter radius as a key physical parameter in physics-informed predictive models for lithium-ion battery failure.

% \begin{figure}[H]
%     \centering
%     \begin{subfigure}[t]{0.48\columnwidth}
%         \centering
%         \includegraphics[width=\linewidth]{f-soc-9m.png}
%         \caption{Peak force dependence on indenter radius and SOC}
%         \label{fig:geom_force}
%     \end{subfigure}
%     \hfill
%     \begin{subfigure}[t]{0.48\columnwidth}
%         \centering
%         \includegraphics[width=\linewidth]{fig_SalC/FailureTime-Distance-5e.png}
%         \caption{Electrical failure time under different contact conditions}
%         \label{fig:geom_failure_time}
%     \end{subfigure}

%     \medskip

%     \begin{subfigure}[t]{0.48\columnwidth}
%         \centering
%         \includegraphics[width=\linewidth]{fig_SalC/Time-intenter-12t.png}
%         \caption{Temperature evolution under different indenter radii}
%         \label{fig:geom_temp_time}
%     \end{subfigure}

%     \caption{Effect of indenter geometry on the coupled response.}
% \end{figure}

% -------------------------------------------------
\subsubsection{State-of-Charge Effect}

Under needling conditions, the mechanical response is highly localized and limited to the direct penetration of the electrode-separator assembly, with negligible global deformation of the cell. The needle mechanically breaches the separator and establishes a direct conductive path between electrodes, acting primarily as a trigger for subsequent failure rather than as a governing factor in energy dissipation or damage evolution.

\begin{figure}[H]
    \centering
    \begin{subfigure}[t]{0.48\columnwidth}
        \centering
        \includegraphics[width=\linewidth]{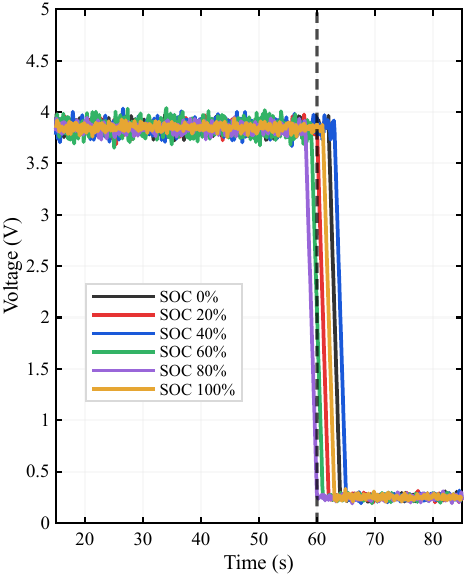}
        \caption{Voltage--time response at different states of charge (SOC)}
        \label{fig:soc_voltage_time}
    \end{subfigure}
    \hfill
    \begin{subfigure}[t]{0.48\columnwidth}
        \centering
        \includegraphics[width=\linewidth]{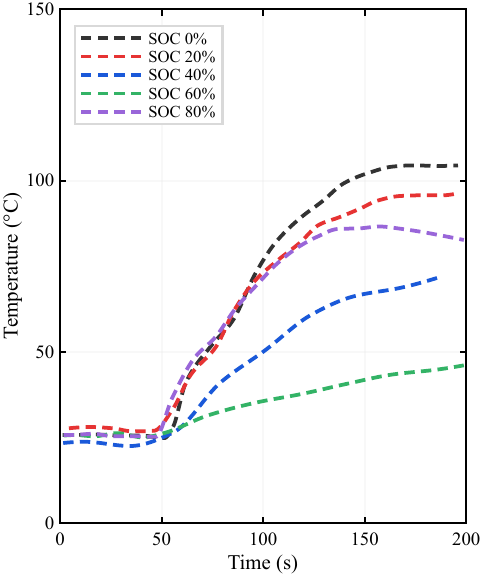}
        \caption{Temperature--time response at different states of charge (SOC)}
        \label{fig:soc_temp_time}
    \end{subfigure}

    \caption{Effect of state of charge (SOC) on the electrical and thermal responses of the cylindrical lithium-ion cell under cylindrical compression.}
    \label{fig:soc_compression_response}
\end{figure}

Following mechanical penetration, the electrical response is characterized by an abrupt voltage collapse, as shown in Fig.~\ref{fig:soc_voltage_time}. Prior to penetration, all cells maintain a near-constant terminal voltage, with higher initial voltage levels observed at increased states of charge (SOC), consistent with open-circuit voltage behavior. Upon penetration, a near-vertical voltage drop occurs for all SOCs, indicating the formation of a hard internal short circuit. Higher-SOC cells exhibit earlier voltage collapse and a larger magnitude of voltage reduction, reflecting the increased availability of electrochemical energy.

The magnitude of voltage collapse increases with state of charge (SOC), reflecting the higher stored electrical potential. This behavior supports a linear scaling between voltage drop and SOC,
\begin{equation}
\Delta V \propto \mathrm{SOC},
\end{equation}
where $\Delta V$ denotes the total voltage reduction at failure. Unlike cylindrical compression, the voltage transition under needling conditions is extremely abrupt, highlighting the localized and direct nature of the short-circuit path created by the needle.

The thermal response, presented in Fig.~\ref{fig:soc_temp_time}, closely follows the electrical failure. At low SOC, the temperature rise remains limited, indicating weak Joule heating due to reduced short-circuit current. As SOC increases, both the rate of temperature rise and the peak temperature increase significantly, with high-SOC cells exhibiting rapid thermal escalation immediately after voltage collapse. The near-simultaneous onset of temperature rise and voltage failure confirms that thermal evolution during needling is electrically driven rather than mechanically induced.

The relationship between peak temperature and state of charge (SOC) follows a logarithmic trend,
\begin{equation}
T_{\mathrm{peak}} = T_{0} + k \ln(\mathrm{SOC}),
\end{equation}
where $T_{0}$ is the baseline temperature prior to failure and $k$ quantifies the sensitivity of thermal escalation to SOC. This functional form captures the strong increase in thermal severity at low-to-moderate SOC and the gradual saturation tendency at very high SOC.

Taken together, the responses shown in Fig.~\ref{fig:soc_compression_response} demonstrate a tightly coupled electro-thermo-mechanical failure mechanism dominated by SOC. Mechanical penetration initiates separator breach, electrical short-circuiting governs energy release, and thermal escalation reflects the magnitude and rate of Joule heating. With increasing SOC, voltage collapse occurs earlier and more abruptly, thermal runaway initiates faster, and peak temperatures increase substantially, identifying SOC as the primary amplifier of failure severity under penetration-type abuse.

A combined severity index integrating force, temperature, voltage drop, and collapse time was defined as
\begin{equation}
S = w_{F}\frac{F_{\mathrm{peak}}}{F_{\mathrm{ref}}}
  + w_{T}\frac{T_{\mathrm{peak}}}{T_{\mathrm{ref}}}
  + w_{V}\frac{\Delta V}{\Delta V_{\mathrm{ref}}}
  + w_{t}\left(\frac{t_{\mathrm{ref}}}{t_{\mathrm{collapse}}}\right),
\end{equation}
where $F_{\mathrm{peak}}$ is the peak indentation force, $T_{\mathrm{peak}}$ is the maximum temperature attained during failure, $\Delta V$ is the total voltage drop at collapse, and $t_{\mathrm{collapse}}$ denotes the time to voltage collapse. The reference quantities $F_{\mathrm{ref}}$, $T_{\mathrm{ref}}$, $\Delta V_{\mathrm{ref}}$, and $t_{\mathrm{ref}}$ are normalization constants, while $w_{F}$, $w_{T}$, $w_{V}$, and $w_{t}$ are weighting coefficients reflecting the relative contribution of each failure metric.

This index exhibits a strong monotonic increase with state of charge (SOC), identifying 80\% SOC as the most hazardous condition.

% \begin{figure}[H]
%     \centering
%     \begin{subfigure}[t]{0.48\columnwidth}
%         \centering
%         \includegraphics[width=\linewidth]{fig_SalC/Voltage-time-SOC-16e.png}
%         \caption{Voltage--time response at different states of charge}
%         \label{fig:soc_voltage_time}
%     \end{subfigure}
%     \hfill
%     \begin{subfigure}[t]{0.48\columnwidth}
%         \centering
%         \includegraphics[width=\linewidth]{failuretime-soc.png}
%         \caption{Failure time as a function of state of charge}
%         \label{fig:soc_failure_time}
%     \end{subfigure}

%     \medskip

%     \begin{subfigure}[t]{0.48\columnwidth}
%         \centering
%         \includegraphics[width=\linewidth]{fig_SalC/temp-time-soc-17.png}
%         \caption{Temperature evolution at elevated states of charge}
%         \label{fig:soc_temp_time_high}
%     \end{subfigure}
%     \hfill
%     \begin{subfigure}[t]{0.48\columnwidth}
%         \centering
%         \includegraphics[width=\linewidth]{fig_SalC/temp-time-soc-11.png}
%         \caption{Temperature evolution across different states of charge}
%         \label{fig:soc_temp_time_all}
%     \end{subfigure}

%     \caption{Effect of state of charge on the coupled mechanical--electrical--thermal response under compressive loading.}
%     \label{fig:soc_response}
% \end{figure}

% ----------------

\subsection{LSTM-Based Temperature Prediction}

\begin{figure}[h!]
\centering
\includegraphics[width=0.48\textwidth]{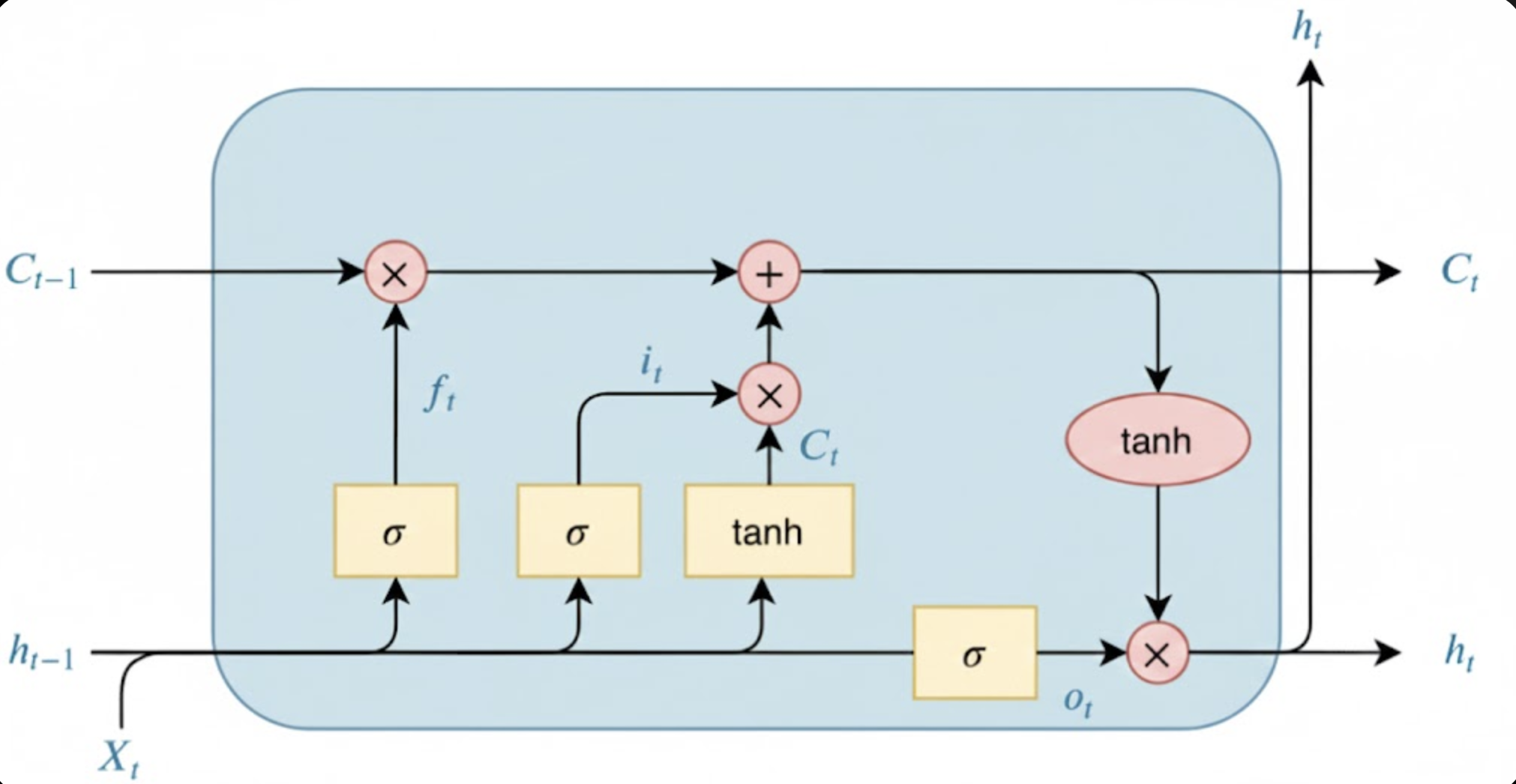}
\caption{Structure of the LSTM network used for temperature prediction.}
\label{fig:lstm_structure}
\end{figure}

The Long Short-Term Memory (LSTM) network serves as the foundational deep learning model used in this study for temporal temperature prediction. LSTMs are specifically designed to capture long-range dependencies through gated mechanisms, namely the input, forget, and output gates, which regulate information flow across time steps. This architecture enables the model to retain relevant thermal dynamics while suppressing noise or transient fluctuations in sensor measurements. The overall structure of the LSTM employed in this work is illustrated in Fig.~\ref{fig:lstm_structure}.

Each LSTM cell captures nonlinear dependencies in temporal sequences through gated mechanisms controlling information flow. Given inputs $\mathbf{x}_t$ and previous hidden states $(\mathbf{h}_{t-1}, \mathbf{c}_{t-1})$, the forward propagation is defined as:
\begin{align}
\mathbf{i}_t &= \sigma(W_i \mathbf{x}_t + U_i \mathbf{h}_{t-1} + \mathbf{b}_i),\\
\mathbf{f}_t &= \sigma(W_f \mathbf{x}_t + U_f \mathbf{h}_{t-1} + \mathbf{b}_f),\\
\mathbf{o}_t &= \sigma(W_o \mathbf{x}_t + U_o \mathbf{h}_{t-1} + \mathbf{b}_o),\\
\tilde{\mathbf{c}}_t &= \tanh(W_c \mathbf{x}_t + U_c \mathbf{h}_{t-1} + \mathbf{b}_c),\\
\mathbf{c}_t &= \mathbf{f}_t \odot \mathbf{c}_{t-1} + \mathbf{i}_t \odot \tilde{\mathbf{c}}_t,\\
\mathbf{h}_t &= \mathbf{o}_t \odot \tanh(\mathbf{c}_t),
\end{align}
where $\sigma$ denotes the sigmoid activation and $\odot$ the Hadamard product. The output $\mathbf{h}_t$ encapsulates temporal information, which is subsequently passed through dense layers to yield temperature forecasts $\hat{T}_t$.
% ================================
%  Figure 2 — Structure of LSTM
% ================================

\subsection{Physics-Informed LSTM-Based Temperature Prediction}

\begin{figure}[h!]
\centering
\includegraphics[width=0.48\textwidth]{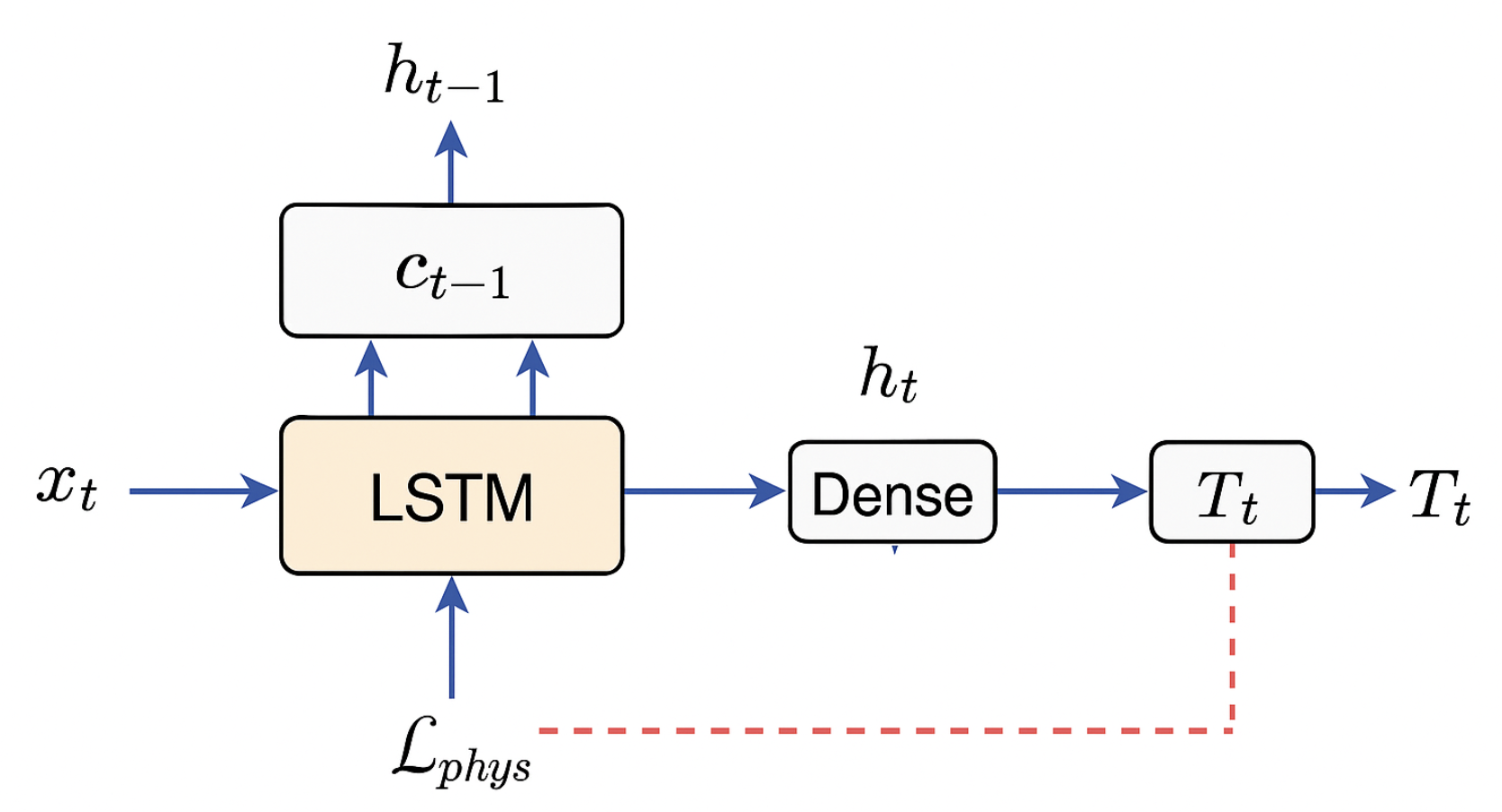}
\caption{Conceptual overview of the proposed PI-LSTM framework, which couples sequential learning with physics-based consistency constraints.}
\label{fig:pi_lstm_arch}
\end{figure}

The proposed Physics-Informed Long Short-Term Memory (PI-LSTM) model enhances the standard LSTM architecture by integrating domain-specific physical knowledge directly into the training process. This is achieved by augmenting the learning objective with a physics-based regularization term, thereby constraining the model outputs to remain consistent with the underlying thermal dynamics. A schematic representation of the overall architecture is provided in Fig.~\ref{fig:pi_lstm_arch}.

To embed physical priors into the model, a regularization term is introduced that penalizes discrepancies from the temporal temperature evolution prescribed by the discretized heat equation. The physics-informed loss is defined as:
\begin{align}
\mathcal{L}_{\text{phys}} 
&= \frac{1}{N} \sum_{t=1}^{N-1}
\left[
\frac{\hat{T}_{t+1}-\hat{T}_{t}}{\Delta t}
-
\alpha \left( 
\hat{T}_{t+1}-2\hat{T}_{t}+\hat{T}_{t-1}
\right)
\right]^{2},
\label{eq:phys_loss}
\end{align}

where $\alpha = k/(\rho c_p)$ denotes the thermal diffusivity, and $\hat{T}_t$ represents the model-predicted temperature at time step $t$. This formulation enforces consistency between the LSTM predictions and the expected thermal diffusion behavior.

The total loss function used for training combines the conventional data-driven objective with the physics-based penalty:
\begin{equation}
\mathcal{L}{\text{total}} = \mathcal{L}{\text{data}} + \lambda \mathcal{L}{\text{phys}},
\label{eq:total_loss}
\end{equation}
where $\mathcal{L}{\text{data}} = \frac{1}{N}\sum_t (T_t - \hat{T}_t)^2$ denotes the mean squared error between measured and predicted temperatures, and $\lambda$ is a hyperparameter governing the influence of the physics-informed constraint. This formulation ensures that the network not only minimizes predictive error but also adheres to physically plausible temperature evolution.

\begin{figure}[h]
\centering
\includegraphics[width=0.48\textwidth]{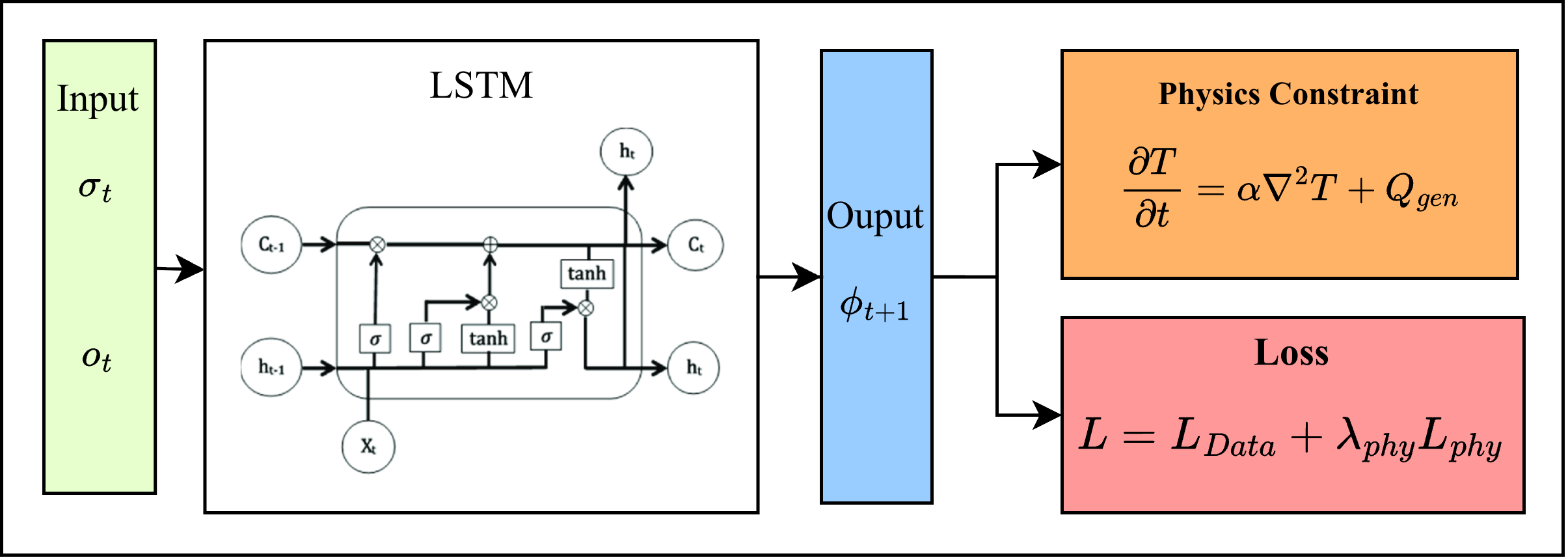}
\caption{Workflow diagram of the PI-LSTM training and inference pipeline.}
\label{fig:flowchart}
\end{figure}

The workflow of the proposed PI-LSTM framework is summarized in Fig.~\ref{fig:flowchart}.

\subsection{Model Performance}

The minimal degradation in accuracy across unseen batteries demonstrates the physical inductive bias imparted by the physics-informed loss term. Unlike purely data-driven models, which often exhibit unstable performance under varying charge-discharge rates, the PI-LSTM maintains consistent prediction fidelity, highlighting its suitability for transfer learning and real-world BMS deployment.

Model performance was quantified using four standard regression metrics:
\begin{align}
\text{MAE} &= \frac{1}{N}\sum_{i=1}^{N} |T_i - \hat{T}_i|,\\
\text{RMSE} &= \sqrt{\frac{1}{N}\sum_{i=1}^{N}(T_i - \hat{T}_i)^2},\\
\text{MAPE} &= \frac{100}{N}\sum_{i=1}^{N}\left|\frac{T_i - \hat{T}_i}{T_i}\right|,\\
R^2 &= 1 - \frac{\sum_i (T_i - \hat{T}_i)^2}{\sum_i (T_i - \bar{T})^2}.
\end{align}
Here, $T_i$ and $\hat{T}_i$ denote actual and predicted temperatures, respectively, and $\bar{T}$ is the mean of actual temperatures. Lower MAE, RMSE, and MAPE values indicate better predictive accuracy, while higher $R^2$ signifies improved explanatory power. Statistical analysis was conducted to assess variance across test batteries.

\section{Results and Discussion}

% Generalization capability was assessed by testing the trained models on Battery~9 and Battery~12, which were excluded from the training phase. As depicted in Table~\ref{tab:generalization}, the PI-LSTM retained strong performance across both test cells, indicating its ability to transfer learned physical relationships across heterogeneous datasets and experimental conditions.

\label{sec:results}
\subsection{Dataset and Data Preprocessing}

% As summarized in Table~\ref{tab:dataset-stats}, thirteen lithium-ion battery datasets (Batt-1 to Batt-13) were employed, each comprising time, state-of-charge (SOC), voltage, current, mechanical force, linear speed, and surface temperature. The datasets were collected under varied ambient conditions, current rates, and mechanical stresses to ensure broad operating diversity. Batteries Batt-1–8, 10–11, and 13 were used for training, while Batt-9 and Batt-12 served exclusively for testing to assess generalization.

\begin{figure}[h]
\centering
\includegraphics[width=0.48\textwidth]{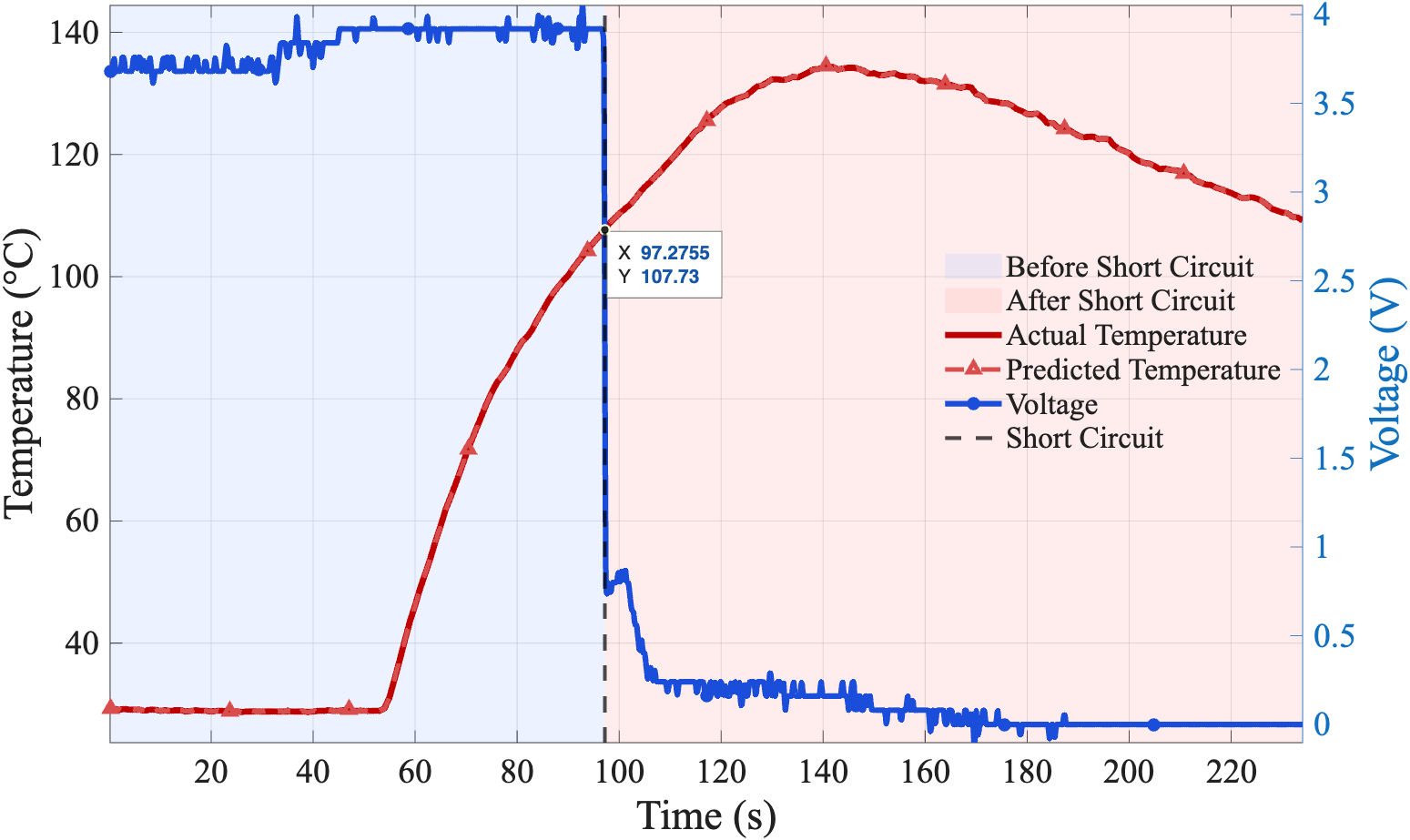}
\caption{Data Analysis}
\label{fig:data_analysis1}
\end{figure}

Fig.~\ref{fig:data_analysis1} illustrates the temporal evolution of temperature and voltage during the short-circuit (SC) event. The plot highlights two distinct operational regions: the pre-SC phase (shaded in blue) where both temperature and voltage remain stable, and the post-SC phase (shaded in red) where a rapid thermal escalation is observed. The actual temperature rises sharply following the SC trigger and continues increasing until reaching its peak, while the PI-LSTM predicted temperature closely tracks this trend with high fidelity. Simultaneously, the voltage exhibits a steep drop immediately after the SC event, eventually stabilizing near zero as the cell undergoes severe degradation. This figure demonstrates the model's capability to accurately capture temperature dynamics during extreme battery stress conditions.

% \begin{figure}[h]
% \centering
% \includegraphics[width=0.48\textwidth]{figure13/Figure_9_dataanalysis.png}
% \caption{Data Analysis of Battery No. 10}
% \label{fig:data_analysis}
% \end{figure}

Prior to modeling, missing values were imputed using forward fill, and categorical entries (e.g., short-circuit indicators) were encoded numerically. All variables were normalized using min-max scaling to the range $[0,1]$. Sequential data windows of fixed length $n=50$ time steps were constructed for temporal learning, following:
\begin{equation}
\mathbf{X}_{i} = [\mathbf{x}_i, \mathbf{x}_{i+1}, \ldots, \mathbf{x}_{i+n-1}], 
\qquad
y_i = T_{i+n}.
\label{eq:sequence_definition}
\end{equation}

This sliding-window method preserves temporal continuity and enables the model to capture both transient and steady-state behaviors.

\begin{figure}[H]
\centering
\includegraphics[width=0.48\textwidth]{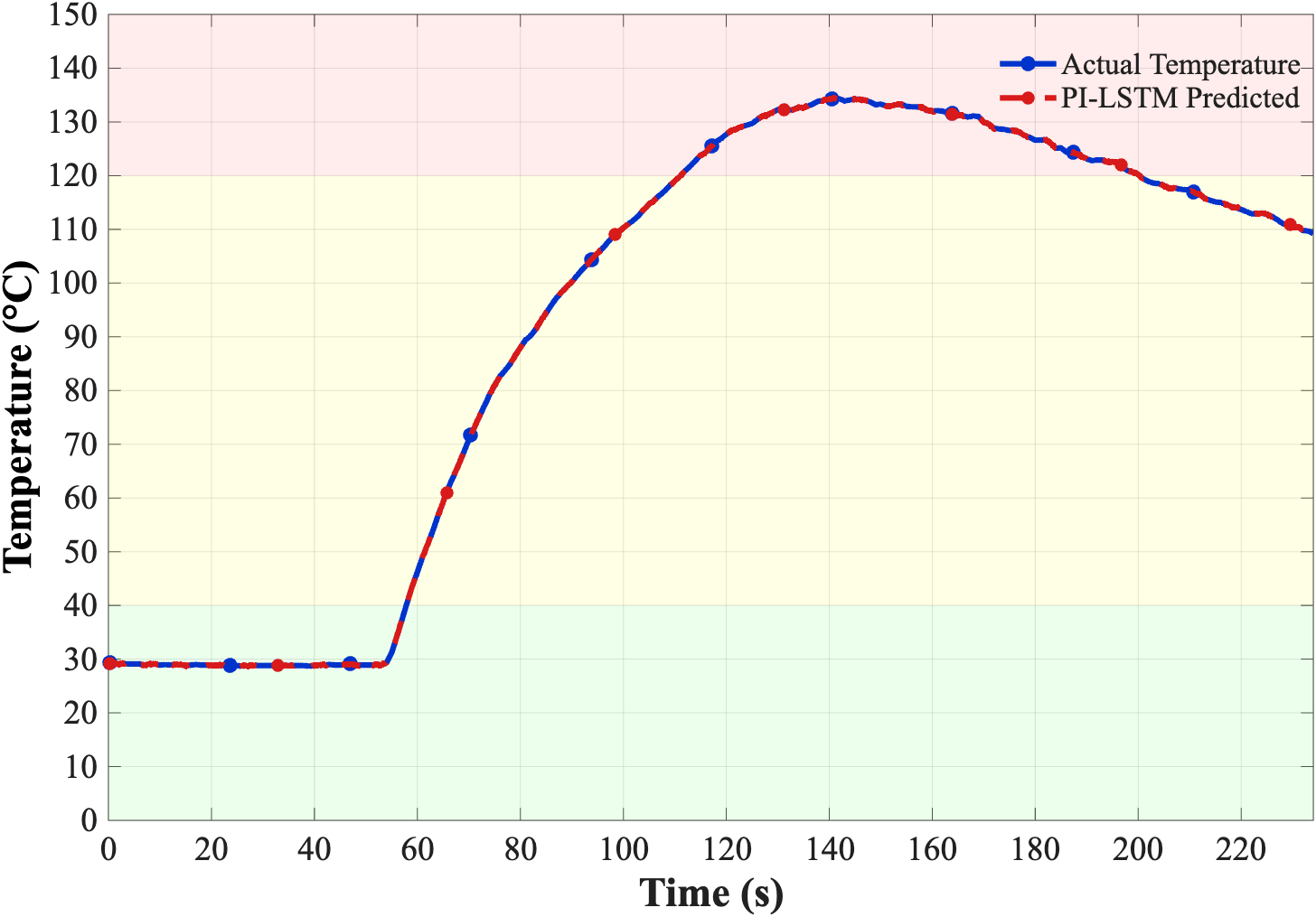}
\caption{Actual vs. PI-LSTM predicted temperature profile for Battery 9.}
\label{fig:actual_vs_predicted}
\end{figure}

Fig.~\ref{fig:actual_vs_predicted} illustrates the comparison between the actual temperature profile and the PI-LSTM predicted response for Battery~9. The plot shows a close alignment between the measured and predicted temperatures across the entire operating period, including the initial steady region, the rapid heating phase, and the peak temperature zone. The shaded background regions visually represent different temperature bands, highlighting the model's ability to accurately track transitions across these zones.

\begin{figure}[H]
\centering
\includegraphics[width=0.48\textwidth]{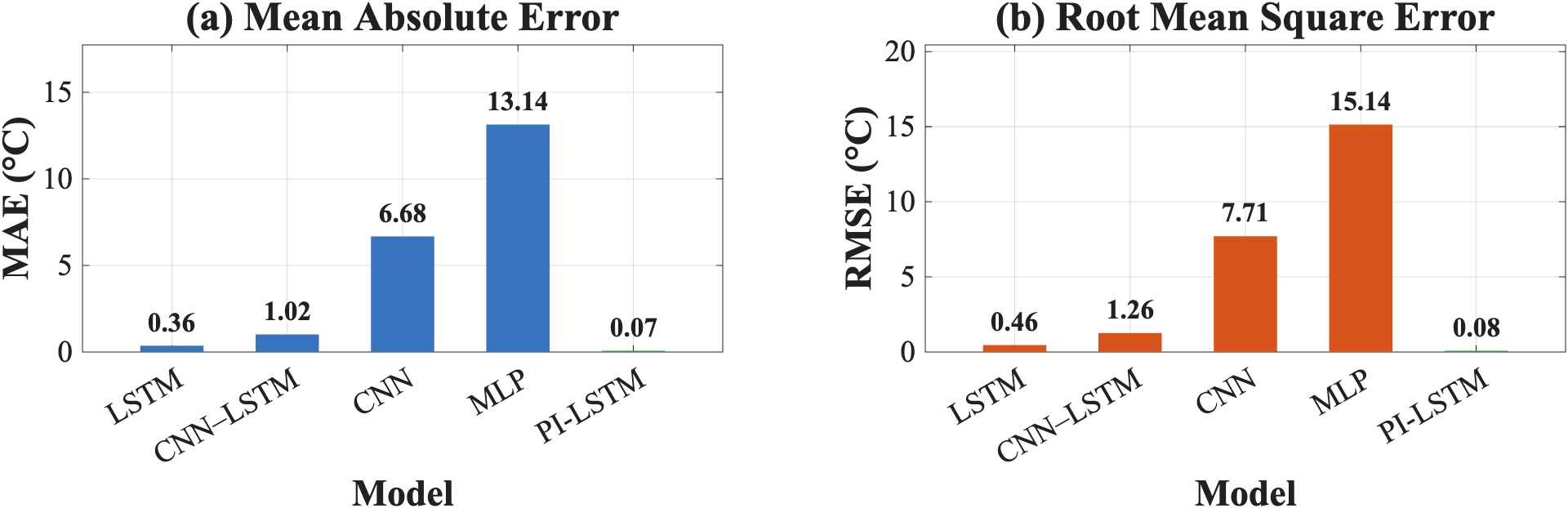}
\caption{Comparison of prediction error metrics (MAE and RMSE) across different models.}
\label{fig:error-metric}
\end{figure}

Fig.~\ref{fig:error-metric} presents a comparative evaluation of different neural network models using two performance indicators: Mean Absolute Error (MAE) and Root Mean Square Error (RMSE). Subfigure~(a) shows that the PI-LSTM achieves the lowest MAE among all models, significantly outperforming LSTM, CNN--LSTM, CNN, and MLP. Similarly, subfigure~(b) demonstrates that the PI-LSTM yields the smallest RMSE, confirming its superior predictive accuracy and robustness relative to the other architectures.

Table~\ref{tab:comparison} summarizes the forecasting performance of various neural network models using MAE, RMSE, and $R^2$ metrics. The results clearly show that the proposed PI-LSTM achieves the best accuracy, with significantly lower error values compared to LSTM, CNN--LSTM, CNN, and MLP models.

\begin{table}[h]
\caption{Performance comparison of temperature-forecasting models.}
\centering
\begin{tabular}{lccc}
\hline
\textbf{Model} & \textbf{MAE ($^\circ$C)} & \textbf{RMSE ($^\circ$C)} & \textbf{$R^2$} \\
\hline
LSTM & 0.3602 & 0.4573 & 0.9999 \\
CNN--LSTM & 1.0188 & 1.2641 & 0.9990 \\
CNN & 6.6788 & 7.7066 & 0.9621 \\
MLP & 13.1394 & 15.1444 & 0.8538 \\
\textbf{PI-LSTM (Proposed)} & \textbf{0.0674} & \textbf{0.0828} & \textbf{0.9998} \\
\hline
\end{tabular}
\label{tab:comparison}
\end{table}

% ================================
% Figure 1 — Experimental Setup
% ================================
% \begin{figure}[H]
% \centering
% \includegraphics[width=0.48\textwidth]{experimental_setup.png}
% \caption{Experimental setup for mechanical abuse and thermal runaway data collection.}
% \label{fig:exp_setup}
% \end{figure}

\subsection{Training of PI-LSTM Model}

Table~\ref{tab:dataset-stats} presents a comprehensive statistical overview of all input features used for training the PI-LSTM model. It includes minimum, maximum, and mean values of time, state-of-charge (SOC), mechanical speed, applied force, voltage, and surface temperature. These statistics illustrate the broad operational diversity captured in the dataset, ranging from low-SOC, low-temperature conditions to high-stress scenarios. The presence of a short-circuit indicator further helps distinguish normal operating phases from TR-triggering events, ensuring that the model learns from both nominal and extreme battery states.

\begin{table}[htbp]
    \centering
    \caption{Statistical summary of the features used for thermal runaway (TR) prediction.}
    \label{tab:dataset-stats}
    \begin{tabular}{lcccc}
        \hline
        \textbf{Feature} & \textbf{Unit} & \textbf{Min} & \textbf{Max} & \textbf{Mean} \\
        \hline
        Time & s & 0 & 3600 & 1800 \\
        SOC & \% & 0 & 100 & 55.2 \\
        Speed & mm/min & 0.5 & 5.0 & 2.8 \\
        Force & kN & 0.2 & 10.0 & 5.1 \\
        Voltage & V & 2.5 & 4.2 & 3.6 \\
        Temperature & $^{\circ}$C & 25.0 & 140.0 & 55.4 \\
        Short Circuit & Yes/No & -- & -- & -- \\
        \hline
    \end{tabular}
\end{table}

The dataset used in this study was divided into separate training and testing subsets to enable robust model development and unbiased performance evaluation. A total of 14,569 datapoints were allocated for training, while 3,642 datapoints were reserved for testing. This split ensures sufficient data coverage for learning the underlying patterns during training while maintaining an independent test set to assess the generalization capability of the proposed model.

% \begin{table}[htbp]
% \caption{Dataset split by number of datapoints for training and testing.}
% \label{tab:dataset-datapoints}
% \centering
% \begin{tabular}{lcc}
% \toprule
% Split & Training & Testing \\
% \midrule
% No. of datapoints & 14,569 & 3,642 \\
% \bottomrule
% \end{tabular}
% \end{table}

% \begin{table}[htbp]
% \caption{Dataset split by number of battery tests for training, validation, and testing.}
% \label{tab:dataset-tests}
% \centering
% \begin{tabular}{lccc}
% \toprule
% Split & Train & Validation & Test \\
% \midrule
% No. of tests & 9 & 3 & 1 \\
% \bottomrule
% \end{tabular}
% \end{table}

The Table~\ref{tab:dataset-tests1} provides a detailed breakdown of the test dataset used to assess the generalization capability of the PI-LSTM model. It includes 13 batteries subjected to various mechanical abuse scenarios such as cylindrical indentation, spherical indentation, radial compression, axial compression, and nail penetration. For each test, key parameters such as SOC, indentation velocity, penetration depth, indenter position, and indenter radius are listed. This diversity ensures that the evaluation covers a wide range of physical stress conditions, enabling robust validation of the model under realistic and extreme battery failure environments.

\begin{table*}[ht]
\centering
\caption{Experimental conditions and parameters for the test dataset used in PI-LSTM model evaluation.}
\label{tab:dataset-tests1}
\begin{tabular}{lcccccc}
\hline
\textbf{Condition} & \textbf{Serial Number} & \textbf{SOC} & \textbf{Velocity (mm/min)} & \textbf{Depth (mm)} & \textbf{Position (mm)} & \textbf{Indenter Radius (mm)} \\
\hline
\multirow{5}{*}{Cylindrical indenter} 
& Batt-1,2 & 0.7,1 & 10 & 13 & 30 & 5 \\
& Batt-3 & 0.4 & 20 & 13 & 30 & 5 \\
& Batt-4 & 0.2 & 10 & 13 & 10 & 5 \\
& Batt-5 & 0.2 & 10 & 13 & 55 & 10 \\
& Batt-6 (deg45$^\circ$) & 0.3 & 10 & 13 & 30 & 5 \\
\hline
Spherical indenter & Batt-7 & 0.5 & 10 & 13 & 30 & 5 \\
\hline
Radial compression & Batt-8 & 0.2 & 2 & 15 & Radial & 100 \\
\hline
Axial compression & Batt-9 & 0.5 & 10 & 20 & Axial & 100 \\
\hline
\multirow{4}{*}{Nail penetration}
& Batt-10 & 0.4 & 20 & 13 & 10 & 2.5 \\
& Batt-11 & 0.4 & 20 & 13 & 55 & 2.5 \\
& Batt-12 & 0.4 & 20 & 16 & 30 & 2.5 \\
& Batt-13 & 0.3 & 20 & 13 & 30 & 2.5 \\
\hline
\end{tabular}
\end{table*}

\begin{figure}[h!]
\centering
\includegraphics[width=0.5\textwidth]{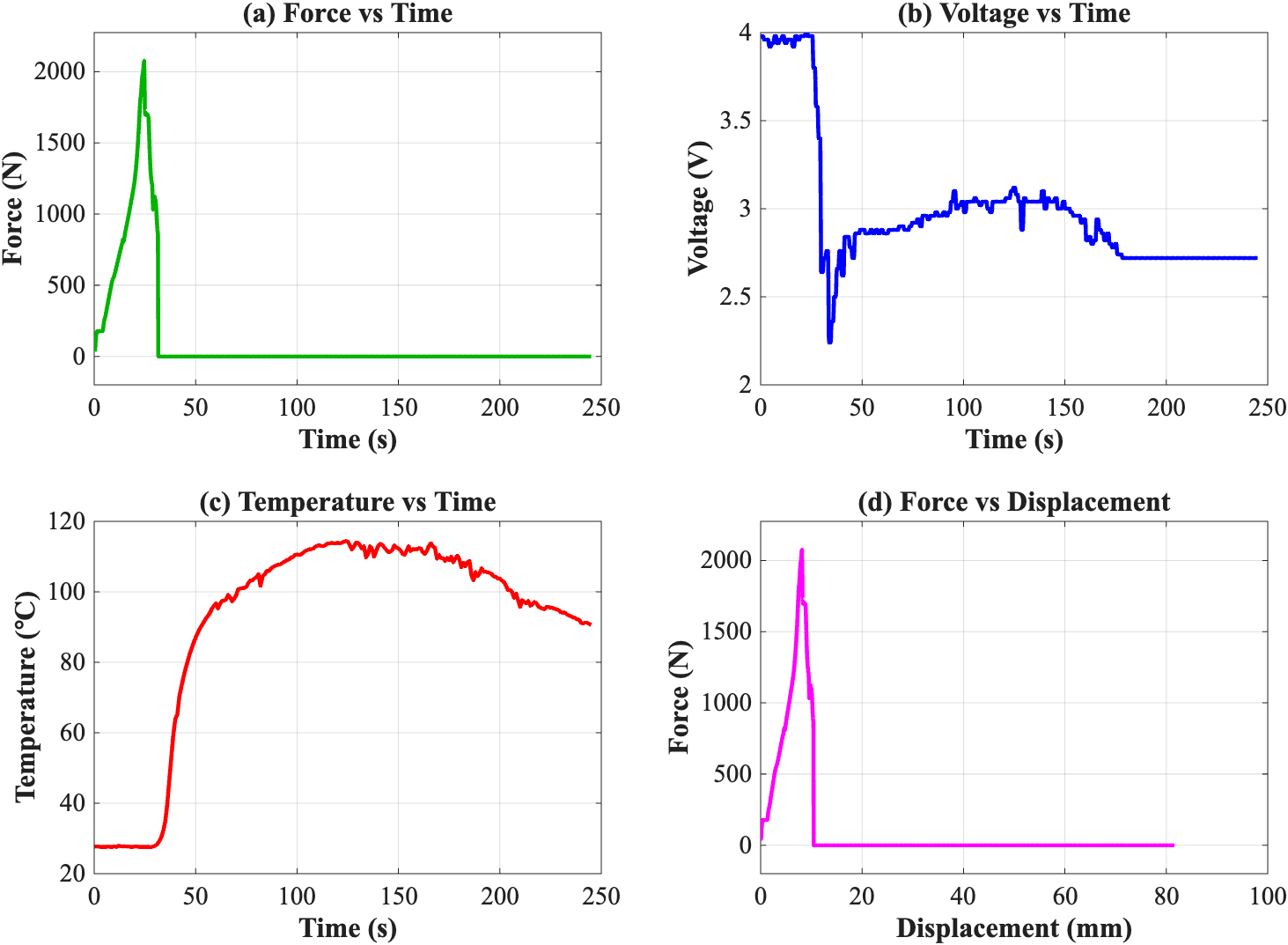}
\caption{Electro-thermal-mechanical response of a lithium-ion battery during a nail penetration test, showing (a) penetration force, (b) voltage, (c) surface temperature, and (d) force-displacement relationship.}
\label{fig:data_analysis}
\end{figure}

This Fig.~\ref{fig:data_analysis} presents a comprehensive dataset from a battery abuse test, likely a nail penetration experiment, showing the temporal evolution of key parameters. Subplot (a) shows the mechanical Force applied over Time, which appears to spike and then drop, indicating the moment of penetration. Subplot (b) tracks the Voltage drop over Time, a clear signature of an internal short circuit. Subplot (c) records the subsequent Temperature rise over Time, capturing the onset of thermal runaway. Finally, subplot (d) provides a mechanical characterization by plotting Force against Displacement, illustrating the battery's structural resistance before failure.

% Figure 2: Description and Figure

\begin{figure}[h!]
\centering
\includegraphics[width=1.0\linewidth]{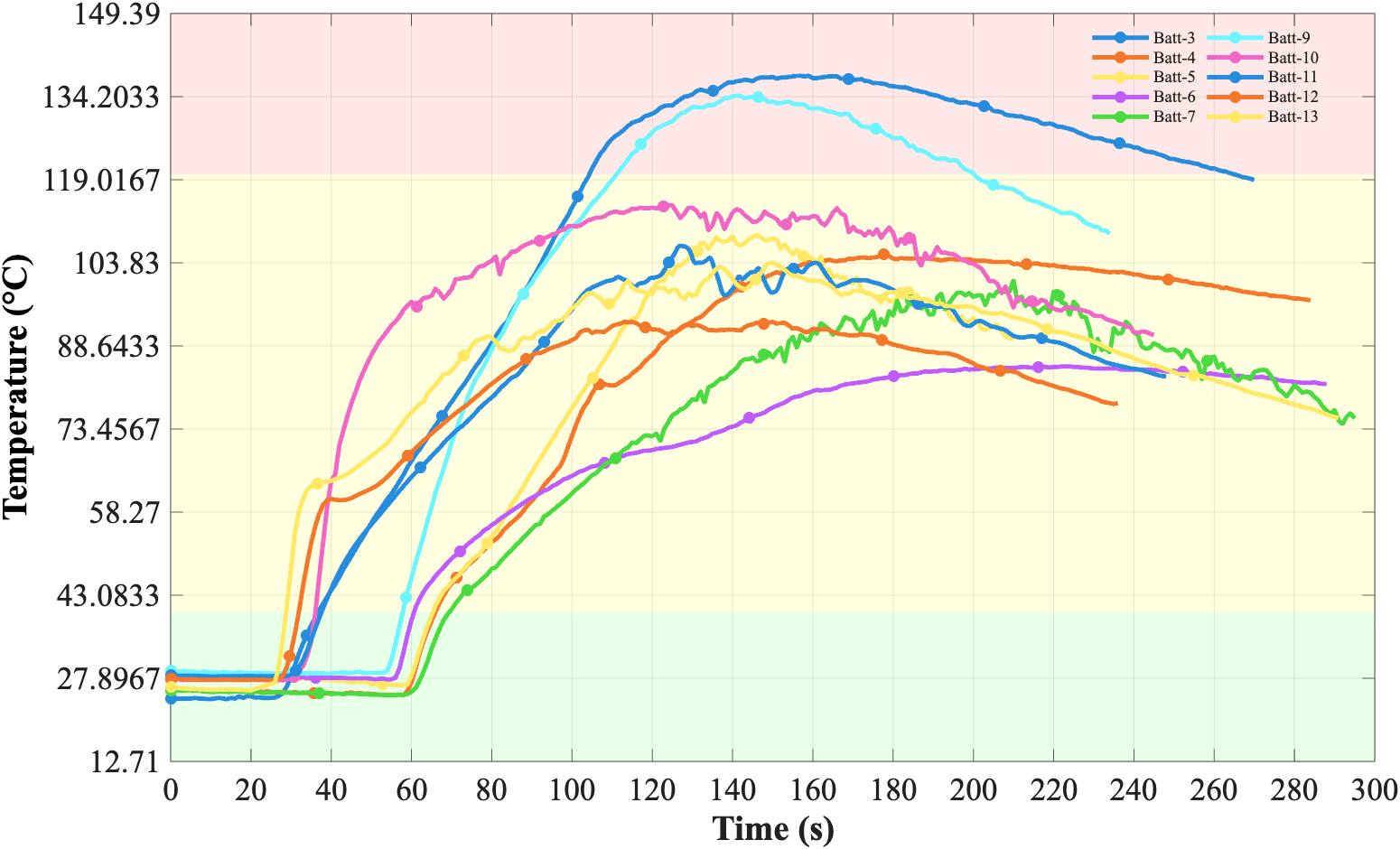}
\caption{Temperature evolution across multiple battery tests used for PI-LSTM model training and evaluation.}
\label{fig:temperature_profiles}
\end{figure}

As referenced in the analysis of thermal data in Fig.~\ref{fig:temperature_profiles}, this figure displays multiple Temperature profiles over Time for a series of battery tests. The overlapping curves, each representing a different test, are typically used to train and validate data-driven models, such as the mentioned PI-LSTM, for predicting thermal behavior. The consistent format allows for comparative analysis of thermal runaway characteristics across different experimental conditions.

% Figure 3: Description and Figure

\begin{figure}[h!]
\centering
\includegraphics[width=0.45\textwidth]{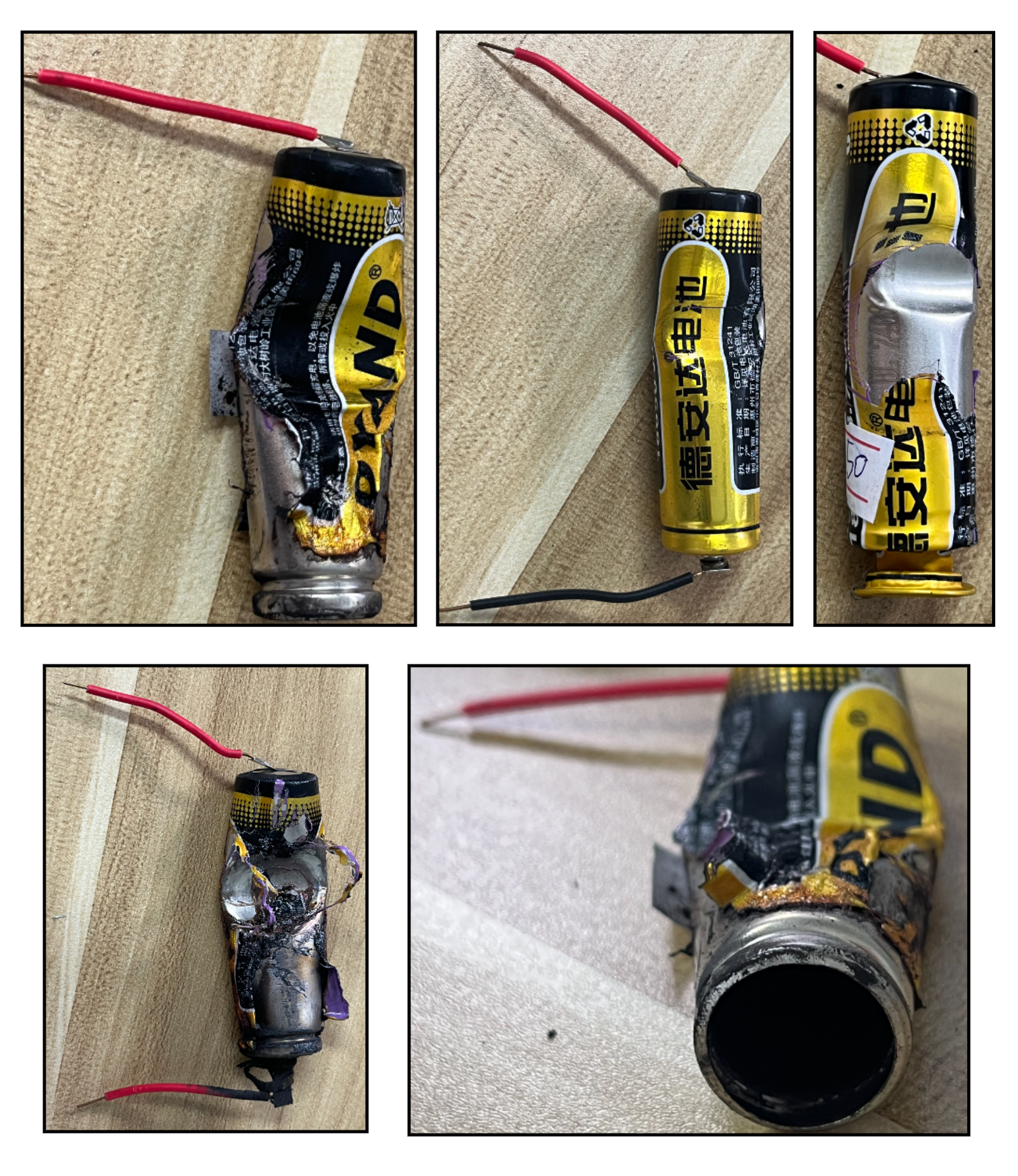}
\caption{Post-test battery damage after thermal runaway experiments.}
\label{fig:damage}
\end{figure}

This Fig.~\ref{fig:damage} provides visual evidence of the physical damage resulting from the severe thermal events. The battery casing is visibly ruptured, deformed, and discolored, with ejected material, confirming the violent nature of the failure. The extent of damage visible here correlates with the extreme temperatures recorded, and demonstrating the destructive potential of thermal runaway in lithium-ion batteries.

Table~\ref{tab:generalization} reports the cross-battery generalization performance of multiple deep learning architectures, including LSTM, CNN--LSTM, CNN, MLP, and the proposed PI-LSTM evaluated on two unseen batteries (Battery 9 and Battery 12). Performance is quantified using mean absolute error (MAE), root mean squared error (RMSE), and the coefficient of determination ($R^2$). The proposed PI-LSTM model significantly outperforms all baseline models, achieving the lowest prediction errors and near-perfect $R^2$ values, demonstrating its superior ability to capture thermal dynamics and generalize across diverse battery types and operating conditions.

\begin{table}[h]
\centering
\caption{Cross-battery generalization performance of different deep learning models on unseen test batteries.}
\resizebox{\columnwidth}{!}{%
\begin{tabular}{lcccc}
\hline
\textbf{Model} & \textbf{Test Battery} & \textbf{MAE ($^\circ$C)} & \textbf{RMSE ($^\circ$C)} & \textbf{$R^2$} \\
\hline
LSTM           & Battery 9  & 0.3602  & 0.4573  & 0.9999 \\
               & Battery 12 & 0.4234  & 0.4525  & 0.9994 \\
CNN--LSTM      & Battery 9  & 1.0188  & 1.2641  & 0.9990 \\
               & Battery 12 & 0.5796  & 0.6855  & 0.9987 \\
CNN            & Battery 9  & 6.6788  & 7.7066  & 0.9621 \\
               & Battery 12 & 2.9941  & 3.2760  & 0.9704 \\
MLP            & Battery 9  & 13.1394 & 15.1444 & 0.8538 \\
               & Battery 12 & 9.3107  & 9.9928  & 0.7248 \\
\textbf{PI-LSTM (Proposed)} & Battery 9  & \textbf{0.0674} & \textbf{0.0828} & \textbf{0.9998} \\
               & Battery 12 & \textbf{0.0503} & \textbf{0.0754} & \textbf{0.9997} \\
\hline
\end{tabular}%
}
\label{tab:generalization}
\end{table}

Experiments use the abuse dataset comprising 13 cells subjected to indentation, compression, and nail penetration across SOC levels (0--100\%). Measurements include synchronized $T$, $V$, and SOC at 0.2~s intervals, with TR onset verified via thermal imaging.

We partition the dataset into disjoint splits: 70\% train, 20\% validation, and 10\% test, ensuring no overlap in experimental conditions. Input sequences span 50 steps (10~s) with forecast horizon $H=30$ (6~s). Features are standardized via z-score normalization.

The UPIT is configured with 3 encoder layers, 4 attention heads, and embedding dimension $d_{\mathrm{model}}=64$. Loss weights are tuned to $(\alpha,\beta,\gamma,\delta)=(0.3,0.4,0.2,0.1)$. Training uses Adam (lr=$10^{-4}$, batch size 32) with early stopping on validation loss. MC dropout is repeated $S=50$ times at inference. On an automotive-grade NVIDIA Drive AGX, inference latency is below 100~ms, satisfying real-time deployment.

% \subsection{Generalization to Unseen Batteries}

\subsection{Thermal Runaway (TR) Warning Strategy}

Based on the Pi-LSTM model proposed in this study, a multilevel thermal runaway (TR) warning strategy has been developed. When the cell temperature exceeds 130~$^\circ$C, large-scale decomposition of the solid electrolyte interface (SEI) occurs~\cite{Feng2020_MitigatingTR}. This process initiates severe side reactions that rapidly increase the battery temperature, thereby elevating the risk of fire or explosion. 

To address this issue, a three-level warning framework is established by integrating the mechanical abuse characteristics of lithium-ion batteries (LIBs) with the temperature threshold corresponding to large-scale SEI decomposition. Alarm level~1 corresponds to external mechanical collision, alarm level~2 is associated with the occurrence of internal short circuits (ISC), and alarm level~3 is triggered when the battery temperature reaches 130~$^\circ$C. 

The TR warning levels are determined based on the Pi-LSTM model's predicted battery failure state and temperature outputs. The results of the TR warning performance under various conditions are illustrated in Fig.~20, showing (a) Batt-1 to Batt-4 and (b) Batt-8 to Batt-10. Experimental validation indicates that the proposed strategy can successfully issue level~3 warnings prior to the onset of thermal runaway, confirming its reliability under different mechanical abuse conditions. In particular, Batt-9 and Batt-12, used as test samples, both triggered level~3 warnings with actual triggering temperatures of 126.78~$^\circ$C and 125.7~$^\circ$C, respectively. 

\begin{figure}[H]
\centering
\includegraphics[width=0.48\textwidth]{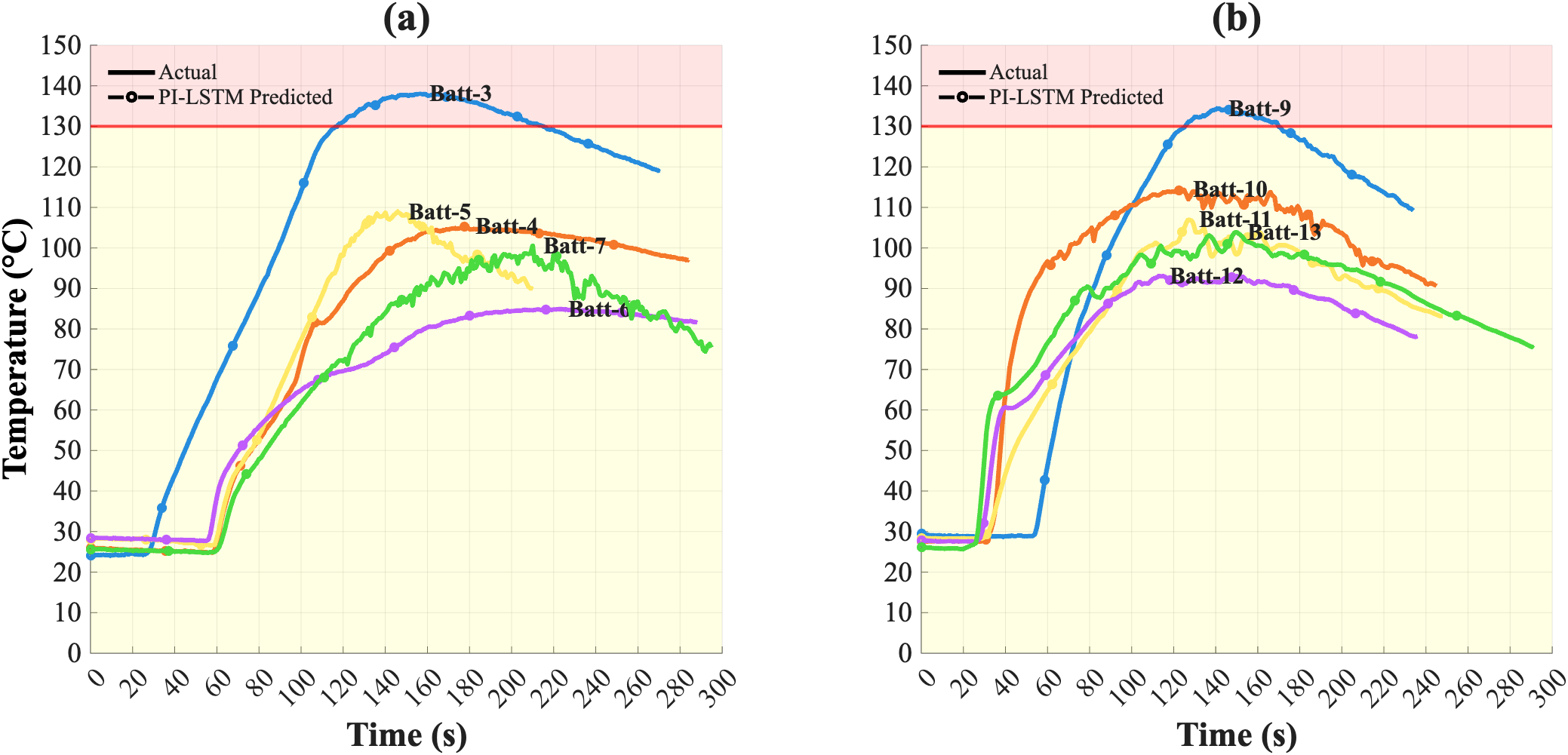}
\caption{Thermal Runaway (TR) warning results}
\label{fig:tr_warning}
\end{figure}

Overall, the multilevel early-warning strategy based on the Pi-LSTM model demonstrates effective early detection of potential thermal hazards, enabling timely preventive measures and improved safety management. Since all datasets correspond to TR events caused by mechanical abuse, alarm level~1 is not shown in Fig.~\ref{fig:tr_warning}.

% =====================================
% Figure 7 — TR Warning Results (Ref)
% =====================================

\section{Conclusion}
\label{sec:conclusion}

This study introduced a Physics-Informed Long Short-Term Memory (PI-LSTM) framework for predicting the thermal behavior of lithium-ion batteries under diverse mechanical abuse and operating conditions. By embedding the heat diffusion equation into the LSTM loss function, the proposed model enforces thermodynamic consistency while retaining the flexibility of data-driven learning. The framework was trained and evaluated using an extensive dataset comprising 13 different battery cells subjected to multiple abuse scenarios, including cylindrical indentation, spherical indentation, radial compression, axial compression, and nail penetration, across state-of-charge levels ranging from 20\% to 100\%. Comprehensive benchmarking against state-of-the-art models---including LSTM, CNN--LSTM, CNN, and MLP architectures---demonstrated that the PI-LSTM consistently delivered superior predictive accuracy and physical coherence. The proposed model achieved up to 81\% lower RMSE, effectively reduced non-physical oscillations, and exhibited improved generalization across varying cell types, mechanical loading conditions, and SOC levels. Overall, the results highlight the potential of physics-informed deep learning as a reliable and robust computational tool for modeling battery thermal response during mechanical abuse, offering significant advantages for safety assessment, early warning systems, and the design of next-generation battery management solutions.

 Future work will extend this approach to multidimensional heat transfer modeling and hybrid electrochemical-thermal forecasting, advancing the development of safer and more intelligent energy storage systems.

\section*{Acknowledgment}
The authors gratefully acknowledge the technical support and research facilities provided by their respective institutions. 
The authors also thank the anonymous reviewers for their insightful comments and constructive suggestions that helped improve the quality of this manuscript.

\section*{Funding}
This work did not receive any specific grant from funding agencies in the public, commercial, or not-for-profit sectors.

\section*{Author Contributions}
All authors contributed to the conception, design, analysis, and interpretation of the study. 
Each author participated in drafting and revising the manuscript and approved the final version for submission.

\section*{Data Availability}
The datasets generated or analyzed during this study are available from the corresponding author on reasonable request.

\section*{Ethics Approval}
This research did not involve human participants or animals; therefore, ethical approval was not required.

\section*{Conflict of Interest}
The authors declare that they have no known financial or personal conflicts of interest that could have influenced the work reported in this paper.

\section*{Supplementary Material}
No supplementary materials are associated with this article. All essential data, figures, and results are included within the main text.


\end{document}